\documentclass{article} % For LaTeX2e
\usepackage{iclr2027_conference,times}

\usepackage{amsmath,amsfonts,bm}

\def\eqref#1{equation~\ref{#1}}
\def\1{\bm{1}}

\DeclareMathAlphabet{\mathsfit}{\encodingdefault}{\sfdefault}{m}{sl}
\SetMathAlphabet{\mathsfit}{bold}{\encodingdefault}{\sfdefault}{bx}{n}

\usepackage{hyperref}
\usepackage{url}

\usepackage{bibunits}
\defaultbibliographystyle{iclr2027_conference}
\defaultbibliography{iclr2027_conference}
\usepackage{multirow}
\usepackage{tabularx}
\usepackage{makecell}
\usepackage{pifont}
\usepackage{xcolor}
\usepackage{colortbl} % for \rowcolor / \cellcolor
\usepackage{caption} % for \captionof
\usepackage{bm}
\usepackage{afterpage}
\usepackage{placeins}
\usepackage{graphicx}
\usepackage{booktabs}
\usepackage[table]{xcolor}
\usepackage{adjustbox}
\usepackage{float}
\usepackage{placeins}
\usepackage{cleveref}

\makeatletter
\newcommand{\figcaption}{\def\@captype{figure}\caption}
\makeatother

\usepackage{amsmath}

\crefname{appendix}{Appendix}{Appendices}
\Crefname{appendix}{Appendix}{Appendices}

\newcommand{\cmark}{\textcolor{green!60!black}{\ding{51}}} % green tick
\newcommand{\xmark}{\textcolor{red!70!black}{\ding{55}}}
\newcommand{\methodhead}[1]{\textbf{#1}}

\definecolor{newcolor}{rgb}{0.8, 0.4, 0.1}

\definecolor{newcolor}{rgb}{0.8, 0.4, 0.1}

\title{MAD-HOI: Masked Autoregressive Diffusion for Generating Articulated Hand-Object Interactions from Text}

\author{\parbox{\dimexpr\textwidth-2\tabcolsep}{\centering
Ananya Bal \quad Kartik Sharma \quad Ethan Lai \quad Samyak Tiwari \\[3pt]
Liza Dahiya \quad Chaitanya Chawla \quad Laszlo A. Jeni}}
\iclrfinalcopy % Uncomment for camera-ready version, but NOT for submission.
\begin{document}
\lhead{Preprint}  
\maketitle
\fancyhf{}                          % clear header and footer
\cfoot{\thepage}                    % restore page numbers
\renewcommand{\headrulewidth}{0pt} 
\vspace{-20pt}
\begin{figure}[H]
    \centering
    \includegraphics[width=0.65\columnwidth]{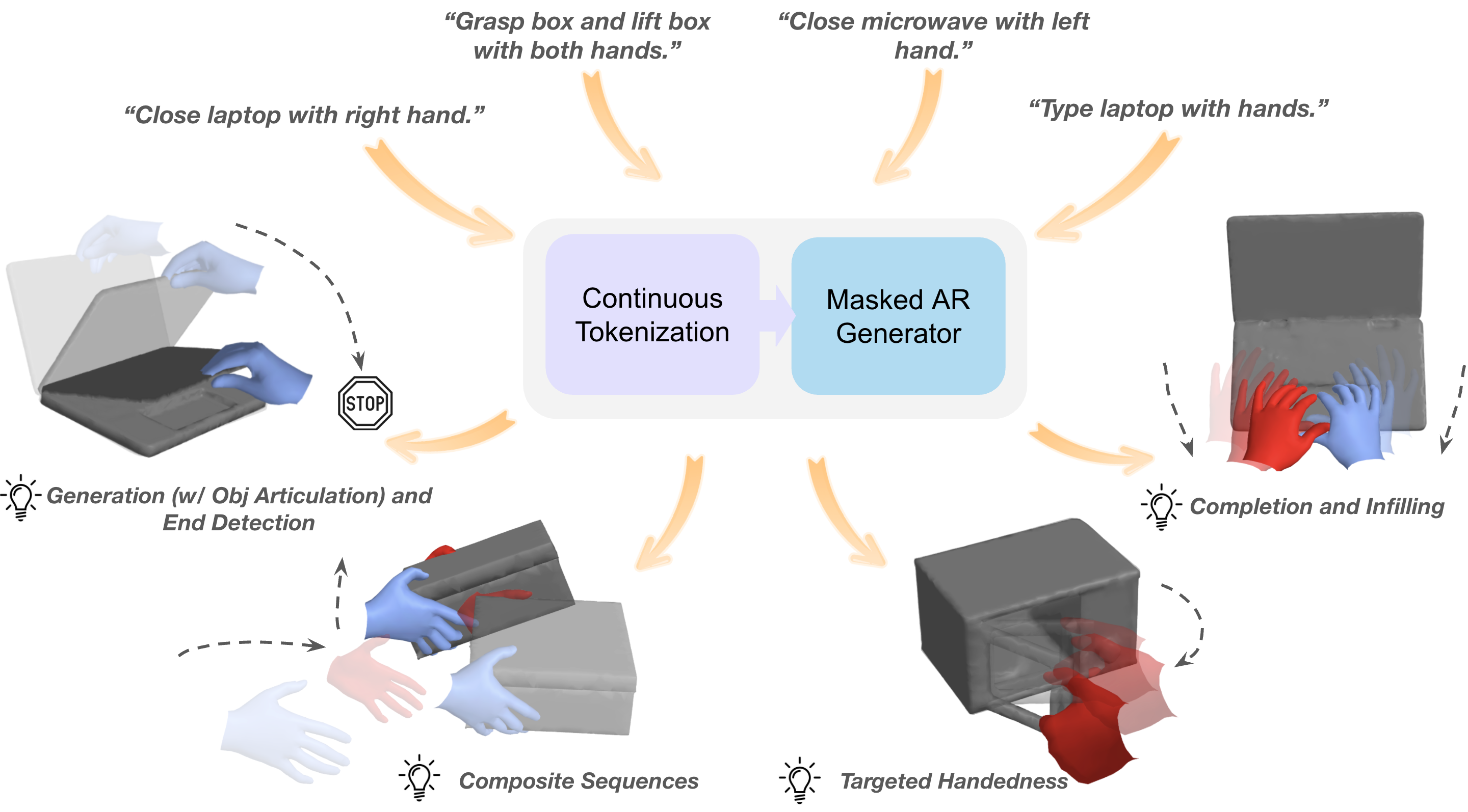}
    \vspace{-4pt}
    \caption{Beyond static grasps and atomic rigid-object motions, MAD-HOI supports articulated and composite interactions, motion completion and infilling, handedness control, and termination through end of motion detection.}
    
    %\textbf{What makes a text-to-motion model for hand-object interaction useful?} We move beyond static grasping poses, rigid objects, and atomic actions: MAD-HOI generates composite motions, completes and infills existing motion, controls which hand acts, articulates objects, and detects its own end of motion.}
    \label{fig:introduction}
    \vspace{-10pt}
\end{figure}

\begin{abstract}
\vspace{-10pt}
Methods for text-based generation of hand–object interaction (HOI) sequences primarily focus on producing smooth, physically plausible trajectories. A truly utilitarian method should additionally support variable-length generation, composite motion sequences, motion completion and infilling, and reliable termination without compromising physical plausibility. Standard diffusion models for HOI generation are typically trained only for text-to-motion generation on atomic motions and require the motion length to be specified \textit{a priori}. Autoregressive (AR) methods provide greater sequence-level flexibility, but commonly depend on discrete motion codes, which can lose contact-sensitive motion detail.  

To address these key limitations, we present a model performing \textbf{M}asked \textbf{A}utoregression with \textbf{D}iffusion for \textbf{HOI} generation \textbf{(MAD-HOI)}. Our method starts by encoding hand and object motions in a continuous latent space while keeping them disentangled to maintain stream-wise control. This is followed by a masked autoregressive transformer to predict context features that condition a flow-matching head. MAD-HOI is capable of motion generation for atomic and composite articulated sequences, conditioned motion completion and infilling, as well as EOM (End of Motion) prediction from a single training objective.

We provide comprehensive evaluations for these capabilities and benchmark our method on the ARCTIC and GRAB datasets. 
Our experiments show that MAD-HOI achieves the strongest text–motion alignment and distributional generation quality against baselines on both ARCTIC and GRAB, while maintaining competitive geometric accuracy and physical plausibility. Code will be made public. 
%\textcolor{red}{Our experiments demonstrate that our method generates more diverse and physically plausible interactions compared to other open-sourced baseline methods.} 
\end{abstract}

% \section{Submission of conference papers to ICLR 2027}

% ICLR requires electronic submissions, processed by
% \url{https://openreview.net/}. See ICLR's website for more instructions.

% If your paper is ultimately accepted, the statement {\tt
%   {\textbackslash}iclrfinalcopy} should be inserted to adjust the
% format to the camera ready requirements.

% The format for the submissions is a variant of the NeurIPS format.
% Please read carefully the instructions below, and follow them
% faithfully.

\vspace{-12pt}
\section{Introduction}
\vspace{-6pt}
Hand manipulation is a fundamental human motor skills ~\citep{cheng2023towards}, and offers a powerful lens into action understanding ~\citep{liu2024taco}. The ability to predict its underlying motions is central to advancing both robotic manipulation \citep{yang2025egovla, singh2025hand} and realism in virtual reality \citep{mangalam2024enhancing}. Studying hand movements for object manipulation through the lens of 3D computer vision is commonly referred to as Hand Object Interaction (HOI). In this paper, we  focus on a specific HOI task - Generation of Hand and Object Motions from Text inputs. In particular, we assume that object geometry and its articulating segments are known, and that the intent of the interaction is provided via language instruction.

Diffusion-based \citep{sohl2015deep} HOI generation methods Text2HOI \citep{cha2024text2hoi}, DiffH2O \citep{christen2024diffh2o}, LatentHOI\citep{li2024latent} produce smooth, realistic motions but are structurally 'atomic': the absence of a strong long-horizon motion prior restricts these approaches to short indivisible movement primitives. Moreover, sequence lengths must be provided by a separate predictor (Text2HOI) or by the user (DiffH2O, LatentHOI). With these methods, generating longer composite sequences from chained instructions is undefined, and trajectories for both hands are generated at all times, with the unused hand masked post-hoc for single-hand prompts.

A complementary line of work - \citet{jiang2023motiongpt, huang2025hoigpt} - offers exactly the missing controls, but pays for them in motion quality. These methods use vector quantization (VQ) to represent motion as discrete tokens and introduce these tokens to an LLM as a `new language'. They gain variable-length generation, completion, and composition through next-token prediction, and termination from the standard EOS (End of Sequence) token. However, quantizing contact-critical motion is destructive: in our tokenizer comparison study (\cref{sec:tokenization}), reconstruction from discrete tokens through a VQ-VAE has much higher penetration and lower contact than with continuous latents from our VAE, and produces visibly spikier jerk profiles (see supplementary). The information lost at tokenization caps the quality of everything generated downstream.

We address this trade-off with a framework that retains the compositional and variable-length benefits of AR modeling while maintaining a continuous motion representation. Our method employs the \textbf{M}asked \textbf{A}utoregressive \textbf{D}iffusion paradigm for text to \textbf{HOI} sequence generation, and is hence named \textbf{MAD-HOI}. We make certain design choices to enable additional capabilities and maximize the expressivity and text-alignment of generated motions.   

We begin by encoding the object, left-hand, and right-hand motions into disentangled continuous latent streams through a Variational Autoencoder (VAE). The VAE is cascaded to condition hand motions on object trajectories and is supervised with contact-aware geometric losses. We then train a bidirectional masked autoregressive transformer to predict conditioning signals for a flow-matching head that generates motion latents. This generator is designed to respect the individual motion streams as separate modalities and predict EOM tokens. Through a single training objective of masked conditional flow matching, and without task-specific heads or post-processing, MAD-HOI learns to respect handedness in text and generate the respective motion streams, complete and infill motion sequences smoothly, generate longer-horizon composite motions and predict EOM tokens eliminating the need for user-specified sequence lengths, as seen in (\cref{tab:capabilities}).

Our contributions are summarized as follows:
\vspace{-8pt}
\begin{itemize}
    \item  We demonstrate that continuous motion latents yield smoother reconstructions than discrete VQ tokens, and propose a cascaded VAE that encodes short temporal sequences into a structured and disentangled latent space.

    \item We show that combining our latent structure with a Masked Autoregressive Diffusion model provides a set of capabilities no existing diffusion HOI method has: varying-length composite generation with learned EOM-based sequence termination, conditioned completion, and infilling while maintaining targeted handedness with genuinely idle hands.

    \item We present comprehensive evaluations and further evaluate text-motion alignment with a protocol wherein the motion evaluator encodes object motion and articulation alongside hand motion, unlike prior protocols that consider hand motion only.
\end{itemize}

\vspace{-10pt}
\section{Related Works}
\vspace{-4pt}

\label{sec:motion-generation}

% \textbf{Full-body motion generation.}
% Diffusion-based methods such as MDM~\citep{tevet2022human}, MotionDiffuse~\citep{zhang2024motiondiffuse}, and MLD~\citep{chen2023executing} model continuous action- or text-conditioned trajectories and generate smooth, diverse motions. In parallel, TM2T~\citep{guo2022tm2t}, T2M-GPT~\citep{zhang2023t2m}, MotionGPT~\citep{jiang2023motiongpt}, and MoMask~\citep{guo2024CVPR} represent motion using discrete VQ tokens, enabling flexible sequence modeling at the cost of quantization. PriorMDM~\citep{shafir2024human} further demonstrates diffusion-based long-duration and spatio-temporal composition, while MARDM~\citep{meng2025rethinking} and MoMADiff~\citep{zhang2025towards} combine continuous diffusion with masked autoregressive generation. 

\textbf{Hand--object interaction generation.} In full-body HOI, CG-HOI~\citep{diller2024cghoi} jointly diffuses contact with human and object motions; OOD-HOI~\citep{zhang2024ood} uses reciprocal diffusion and contact-guided refinement for unseen-object and action generalization; and GenHOI~\citep{li2025genhoi} reconstructs sparse interaction keyframes before synthesizing temporally coherent 4D sequences with contact-aware diffusion.  ARDHOI~\citep{geng2025auto} combines continuous HOI representations with causal autoregressive diffusion. However, its next-token formulation is naturally suited to forward generation, whereas masked bidirectional prediction better suits completion and infilling. These methods lack object and per-hand stream separation for modality control.

Diffusion-based methods for fine-grained hands-only HOI generation typically involve multiple stages, limited sequence-level controllability, and externally supplied motion lengths. Text2HOI~\citep{cha2024text2hoi} first predicts contact maps and then generates motion conditioned on them through diffusion, followed by a penetration-reduction refiner. DiffH2O~\citep{christen2024diffh2o} separates grasping and interaction into two diffusion stages and represents motion in an object-centered coordinate frame. LatentHOI~\citep{li2024latent} similarly adopts an object-centered representation and factorizes generation into latent wrist trajectories and hand-pose decoding. These decompositions can aid unseen-object generalization, but introduce specialized stages and primarily target atomic interactions whose duration must be supplied or separately predicted. They also generate both hand trajectories irrespective of handedness, requiring the unused hand to be suppressed after generation rather than modeled as part of the output distribution. OpenHOI~\citep{zhang2025openhoi} extends a Text2HOI-style generator with multimodal affordance reasoning, physics-based refinement, and a separate post-hoc module for compositing atomic motions.
Autoregressive approaches provide stronger sequence-level flexibility. HOIGPT~\citep{huang2025hoigpt} discretizes hand and object trajectories into VQ-VAE codes and trains a language model over motion and text tokens, naturally supporting variable-length generation, completion, and compositional sequencing. Its capabilities, however, depend on quantized motion representations, which can discard fine-grained kinematic and contact information. MAD-HOI instead retains continuous motion latents while explicitly structuring the sequence into object, left-hand, and right-hand streams, allowing the model to generate asymmetric bimanual interactions, represent inactive hands directly, and predict the termination of each modality without post-hoc masking or an external length predictor.

Concurrent work, HO-Flow~\citep{chen2026hoflow} independently combines continuous temporal latents with masked autoregressive latent diffusion. The method learns an interaction-aware VAE that projects the object point cloud into the local coordinate systems of the MANO kinematic chain, capturing hand-relative geometry alongside global hand and object motion. It then concatenates the resulting hand--object latents into unified temporal tokens and trains a generator similar to ours. While HO-Flow reports improved physical plausibility over conventional diffusion-based baselines, it does not demonstrate capabilities beyond input-length-constrained text-to-motion generation. Moreover, its validation misses out on articulated-object interactions, text--motion alignment and standard distributional metrics. We omit this baseline as the code is not publicly available and their reported results are not directly comparable as we use different datasets and preprocessing protocols---including different processing of GRAB~\citep{taheri2020grab}.

\vspace{-4pt}
\begin{table}[H]
\centering
\tiny
\setlength{\tabcolsep}{6pt}
\renewcommand{\arraystretch}{1.2}

\begin{tabular}{@{}lccccc>{\columncolor{blue!10}}c@{}}
\toprule
\textbf{Capability}
& \methodhead{Text2HOI}
& \methodhead{DiffH2O}
& \methodhead{LatentHOI}
& \methodhead{OpenHOI}
& \methodhead{HOIGPT}
& \methodhead{MAD-HOI}
\\[-0.4ex]
\midrule
Continuous Repr.  & \cmark & \cmark & \cmark & \cmark & \xmark & \cmark \\
Motion Generation & \cmark & \cmark & \cmark & \cmark & \cmark & \cmark \\
Auto-termination (EOM)  & \xmark & \xmark & \xmark & \xmark & \cmark & \cmark \\
Motion Completion & \xmark & \cmark & \cmark & \xmark & \cmark & \cmark \\
Motion Infilling  & \xmark & \xmark & \xmark & \xmark & \cmark & \cmark \\
Composite Motion  & \xmark & \xmark & \xmark & \cmark & \cmark & \cmark \\
Captioning        & \xmark & \xmark & \xmark & \xmark & \cmark & \xmark \\
\bottomrule
\end{tabular}
\vspace{-4pt}
\caption{\textbf{Capability comparison} of MAD-HOI with relevant baselines (determined from their papers and public code.)}
\vspace{-14pt}
\label{tab:capabilities}
\end{table}
% \vspace{-12pt}
% We however, compare our method against all relevant open-sourced baselines.
\vspace{-12pt}
\section{Tokenization: Discrete vs Continuous}
\label{sec:tokenization}
\vspace{-8pt}
Autoregressive modeling operates on a tokenized latent space, and the tokenizer bounds the quality of downstream generations.
VQ-VAE-based methods \citet{jiang2023motiongpt, huang2025hoigpt} encode motion into discrete codebooks and train an LLM to predict the code indices; we argue that discretization loses information HOI cannot afford to lose, as small quantization errors in wrist pose or finger articulation translate directly into penetration, broken contact and jitter.

To demonstrate this, we train our VAE and an equivalent VQ-VAE architecture (similar to \citet{huang2025hoigpt}) and compare their reconstructions, measuring penetration and contact percentages following \citet{zhang2025bimart} and, following \citet{Cho_2025_ICCV}, jerk profiles (the third derivative of position) which exposes high-frequency jitter that mean errors hide; plots presented in the \cref{app:disc_v_cont}.
\textbf{\textit{The VAE reconstructs with penetration in only 2.1\% of frames and contact in 92.2\% frames, against 24.7\% and 77.9\% of the frames respectively for the VQ-VAE. And the jerk profiles show that the VAE tracks the ground truth closely and produces less jerky (jittery) motion, especially when it comes to reconstructing the longer tail of static hands.}}
This motivates continuous motion latents over discrete motion tokens for autoregressive HOI generation.

\vspace{-8pt}

\section{Method}
\begin{figure}[H]
  \centering
  \includegraphics[width=0.9\textwidth]{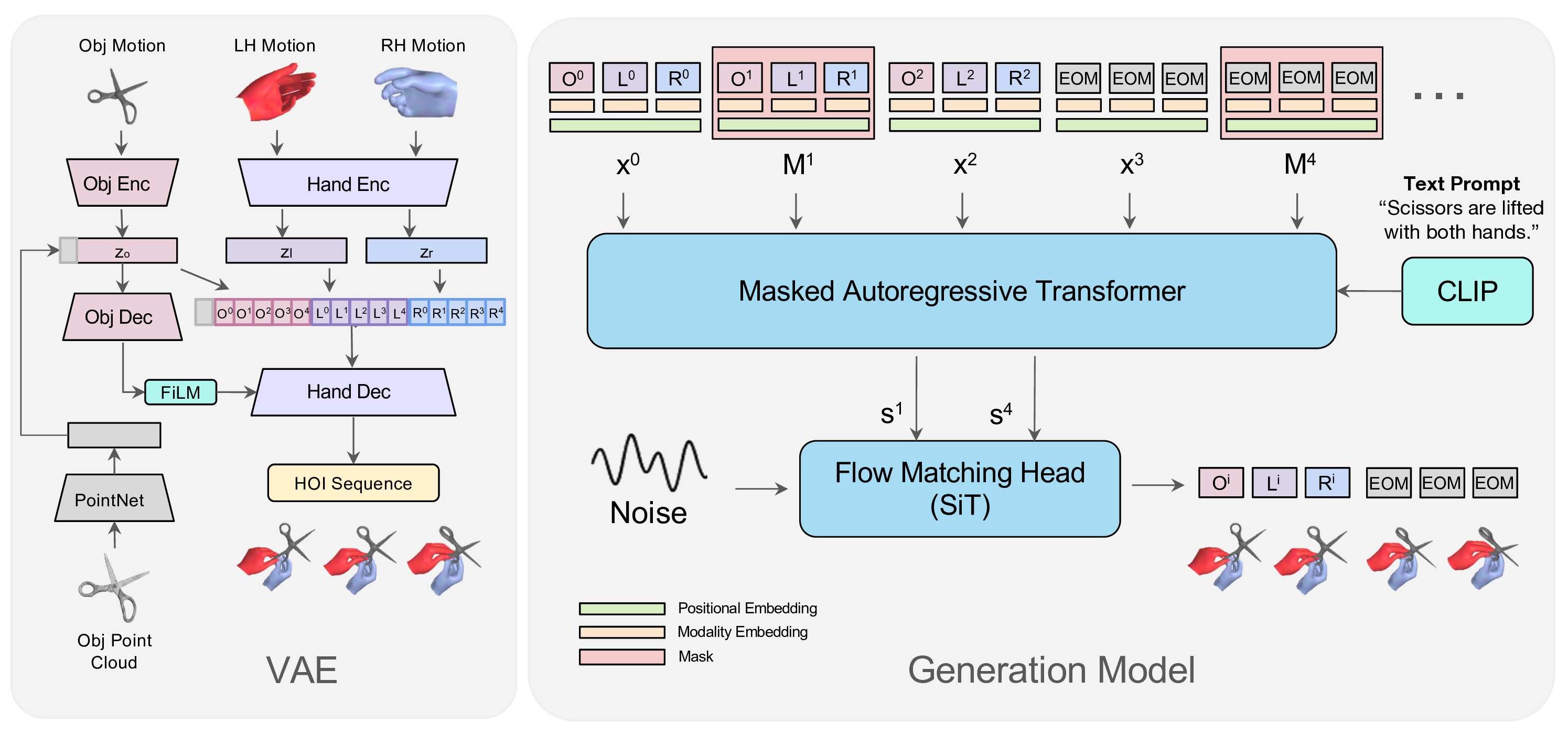}
  \vspace{-8pt}
  \caption{\textbf{MAD-HOI architecture.} Left: Our cascaded VAE builds continuous motion latents for object and hand trajectories. Right: Our generation model consists of a masked autoregressive transformer that predicts conditioning signals for a flow-matching head - generating motion latents for masked tokens, enabling smooth,  text-based HOI generation, completion, infilling and termination.}
  \label{fig:arch}
  \vspace{-12pt}
\end{figure}

\vspace{-6pt}
MAD-HOI first uses a cascaded VAE (left in \cref{fig:arch}) to create motion latents for object and hand trajectories.
These motion latents then undergo masking before being passed to the autoregressive diffusion generator of our framework (right in \cref{fig:arch}), which consists of a masked autoregressive transformer producing conditions for the diffusion head at masked positions.
Our method derives its capabilities from two streams - 1. The bidirectional autoregressive model with iterative masked denoising allows the generator to look at the full unmasked temporal context while generating each motion latent, giving rise to conditioned completion and infilling (depending on the masking). 2. Our VAE with it's disentangled the latent structure enables modality control and stream-wise generation/completion/infilling which improves `handedness'. It further learns `idle hand' latents which enable the generator to get the right context for predicting EOM and therefore, sequence termination.

\vspace{-4pt}

\paragraph{Data:}We focus on motion datasets that capture one-handed and bimanual object interactions in 4D.
GRAB \citep{taheri2020grab} captures full-body sequences with rigid objects, of which we use only the hand interaction data; ARCTIC \citep{fan2023arctic} consists of single-joint articulating objects.
With the preprocessing protocol and text annotations from \citep{cha2024text2hoi}, GRAB has 1.6k motion sequences for 51 objects and ARCTIC 4.6k atomic sequences spanning 11 objects; this difference in trajectories per object prompts us to maintain separate experiments for the two datasets.

The object trajectory is a list of global poses $O = (T_o^t, R_o^t, \alpha_o^t)$, where $T_o$ is the normalized global 3D location of the object center, $R_o$ its 6D rotation, $\alpha_o$ an optional articulation value for articulated objects, and $t$ the time frame.
Hand poses are based on MANO \citep{romero2022embodied}: each hand's trajectory is $H = (\tau_h^t, \phi_h^t)$, with $\tau_h$ the normalized global 3D wrist position and $\phi_h$ the 6D rotations of the wrist and the other 15 MANO joints, plus a binary left/right identifier appended separately.

\vspace{-4pt}

\subsection {Variational Autoencoder} \label{subsec:vae}
\vspace{-6pt}
To enable modality control in generation, our tokenizer must keep the three motion modalities (object, left hand and right hand) as separate  latent streams.
Following prior works \cite{jiang2023motiongpt, meng2025rethinking, huang2025hoigpt}, we use 1D conv + resnet blocks in the encoders and decoders of our cascaded VAE.
We encode the object motion and the hand motions through different branches. We further recognize that hand motions depend heavily on the object motion and therefore design a cascaded network structure that decodes hands conditioned on the decoded object motion.  
\textbf{Object branch:} The object encoder encodes $O$ into a 512-dimensional latent $z_{o}$, and the object decoder receives pointnet features from a canonical object point cloud, which are concatenated with $z_{o}$ to reconstruct object trajectories. 
\textbf{Hand branch: }The hand encoder and decoder maintain one set of weights to separately process two streams of input, $H_{left}$ and $H_{right}$, latents $z_{l}$ and $z_{r}$ for the left and right hands respectively. This enables co-learning and efficiently embeds both hand motion streams into a shared latent space while keeping them differentiated through an identifier value. The object latent $z_{o}$ is concatenated with $z_{l}$ and $z_{r}$ and passed to the hand decoder to jointly decode the full HOI sequence. The hand decoder also applies the reconstructed $O$ as conditions through a FiLM layer \citep{perez2018film}. 

In addition to the motion reconstruction objective, we force our VAE to reconstruct static poses over short horizons as `idle latents'. We do this to have hand inactivity represented by static (or `idle') poses, lending to true `handedness' without post-hoc masking of either hand. See \cref{subsec:ard} for the use. Each encoder predicts a posterior mean and variance, and in the generator training, motion latents are drawn with the reparameterization trick, $z = \mu + \sigma \epsilon$ with $\epsilon \sim \mathcal{N}(0, I)$.
A KL-divergence loss regularizes each posterior toward a unit Gaussian.
%; \kartik{we keep its weight small so that reconstruction dominates}.
This regularization matters downstream: the flow-matching head of our generator regresses velocities toward samples from this latent space, and a bounded, smoothly populated space is a more stable regression target than the unconstrained latents of a deterministic autoencoder (see supplementary for this comparison).

We use a window length of 152 frames with a 4-frames-to-1-latent temporal compression throughout, giving at most 38 latents per modality (see \cref{fig:arch} for the architecture). The latents are 512 dimensional. The VAE is trained to reconstruct HOI sequences with L1 reconstruction loss and three geometric losses (definitions in the supplementary): a contact loss $\mathcal{L}_C$ penalizing hand joints that stray beyond a threshold distance $\phi$ from the object surface, a penetration loss $\mathcal{L}_{pen}$ penalizing hand vertices inside the object mesh, and a distance map loss $\mathcal{L}_{DM}$ \citep{liang2024intergen} aligning predicted and ground-truth joint-to-object distance maps in near-contact regions.
The total loss combines L1 losses over the translation and rotation components of the object and hand trajectories with the geometric losses and the KL term. Weights are detailed in the \cref{app:training_details} and network specification in \cref{app:nw_specs}.
\begin{equation}
    \mathcal{L}=\alpha\mathcal{L}_{rec} + \beta\mathcal{L}_C + \gamma\mathcal{L}_{pen} + \delta\mathcal{L}_{DM} + \lambda\mathcal{L}_{KL}
\end{equation}
% We use $\alpha=1$, $\beta=0.5$, $\gamma=0.5$, $\delta=1$, and $\lambda=$0.5.

\vspace{-6pt}

\subsection {Autoregressive Diffusion } 
\label{subsec:ard}
\vspace{-6pt}
Unlike diffusion models, autoregressive models factorize the conditional likelihood of a sequence, $p(x^{1:n}\mid c)$, into chained conditionals $p(x^i\mid x^{<i},c)$, generating each token conditioned on the previously generated tokens and the input condition (c). Extending diffusion models for AR prediction with an MSE objective over $\epsilon_\theta$ or $x_0$ collapses the generation task into simple regression with gaussian probability, failing to model the chained conditional probabilities that define $p(x^{1:n}|c)$ in true autoregressive generation.
Recent advances in image generation \citep{li2024autoregressive} highlight the promise of continuous autoregressive generation, where logits from an AR model are used as conditioning signals for a continuous sampling network to better approximate underlying data likelihoods with chained probability.

At each autoregressive iteration in MAD-HOI, interaction sequences are encoded into motion latents through the VAE and masked through a cosine schedule \citep{li2024autoregressive}.
The unmasked latents form the visible context $\mathbf{um}$, while the masked latents $\mathbf{m}$ represent the 'tokens' that must be generated under condition $c$.
Generation proceeds over $K$ unmasking iterations, with $\mathbf{m}_j$ denoting the subset of masked latents generated at iteration $j$; see \cref{eq:mar_factorization}
\vspace{-2pt}
{\small
\begin{equation}
p(x'_{1:n} \mid c)
= \prod_{j=1}^{K} p\!\left(\mathbf{m}_j \mid \mathbf{um},\, \mathbf{m}_{<j},\, c\right).
\label{eq:mar_factorization}
\end{equation}}
\vspace{-12pt}
\subsubsection{Masked Autoregressive Transformer}
\vspace{-2pt}
% Every capability we target is a masking pattern, so one masked-generation objective can train them all.
% Generating from scratch masks everything; completing a motion masks the future; infilling masks an interior segment; composing masks everything after a generated prefix; and controlling handedness amounts to generating one hand's stream while the other encodes rest.
% The transformer's job is to turn the visible context into meaningful conditions $s$ for the diffusion head at every masked position.
% For these masking patterns to act as controls, however, the token sequence itself must expose the structure they operate on.

The transformer is tasked with capturing the context of time-variant motion data, and provide meaningful conditions $s$ for the flow matching head. In our case, an interaction sequence encoded with our VAE results in three motion latents,  $z_o, z_l, z_r $ which are all considered as different motion modalities in our framework. We refer to this triplet of latents as $x$ and maintain this structure per timeframe by interleaving latents from the three modalities and attaching one positional encoding to each triplet. The latent from each modality also receives its own learnable \textbf{\textit{`modality embedding'}}.  Since our VAE has a decoupled latent space for the hands, when the generation branch recognizes that a hand token corresponds to an 'idle' hand, it is able to predict latents corresponding to 'idleness' leading to the ability to understand 'hand use'. 
%More details are provided in \cref{sub:training_details}. 
While Text2HOI, DiffH2O and LatentHOI generate trajectories for both hands irrespective of the single-hand text prompt and mask the unused hand for visualizations, our method actually predicts static poses for unused hands - leading to \textbf{\textit{`handedness'.}} See \cref{app:handedness} for validation of handedness and idle or unused hands.
Further, we pad all non-valid frames within our input with learnable \textbf{End of Motion $\langle EOM \rangle$ tokens}, while continuing the interleaved structure. The model learns to associate this token with the long tail of `idle' latents and predict the end of generation - meaning that we do not require an input sequence length for inference. A sample input sequence to the generator is provided in the \cref{app:seq}.

Given the unmasked motion latent sequence $\mathbf{um}$ as previously generated motion latents, the masked autoregressive transformer $g$ produces conditions $s = g(\mathbf{um})$ for the diffusion branch to generate latents at masked positions $m$.
Our transformer uses bidirectional attention \citep{meng2025rethinking} and a single transformer layer \citep{ashish2017attention} with Adaptive Layer Norm \citep{peebles2023dit}.
The text caption is fed into the CLIP encoder and integrated into the transformer's latent space through a linear projection.

\vspace{-6pt}
\subsubsection{Diffusion:}
\vspace{-6pt}
The output $s$ from the Masked Autoregressive Transformer is fed into the flow matching head as conditions.
We call this per-token sampler the diffusion head for continuity with \citet{li2024autoregressive, meng2025rethinking}, though its training objective is flow matching.
This network produces the motion latent $x'^i$ at each masked position $i$ by iterating the reverse step $x'^{\,i}_{t-1} \sim p(x'^{\,i}_{t-1} \mid x'^{\,i}_{t}, t, s_i)$ over timesteps $t$.

During the diffusion head training, $k$ random motion latents are masked with a learnable mask vector following the cosine masking schedule from MoMask \cite{guo2024CVPR}.
The autoregressive model learns to provide accurate signals based on the unmasked latents given the timestep and text condition, and the diffusion head uses this signal to predict a velocity $v(x,t)$.
Following SiT \cite{ma2024Sit}, we corrupt the ground-truth latent $x_0$ along the linear interpolant $x_t = \alpha_t x_0 + \sigma_t \epsilon = (1 - t)x_0 + t\epsilon$, where $\epsilon$ is gaussian noise, $t$ is the continuous timestep, and $\dot{\alpha}_t, \dot{\sigma}_t$ denote the time derivatives of the interpolation coefficients (the full velocity-field definition is provided in the \cref{app:fm_details}).
We apply the following loss over velocity prediction:
\vspace{-2pt}
{\small
\begin{equation}
\mathcal{L}_{\text{GB}}
= \int_0^T 
\mathbb{E}_{\mathbf{v},t}
\!\left[
\big\|
\mathbf{v}_{\theta}(\mathbf{x'}^{\,i}_t \mid t, g(\mathbf{u}_m))
- \dot{\alpha}_t \mathbf{x'}^{\,i}_0
- \dot{\sigma}_t \boldsymbol{\epsilon}
\big\|^2
\right]
\, \mathrm{d}t,
\label{eq:velocity_loss}
\end{equation}}
\vspace{-12pt}

At inference, given previous motion latents $\mathbf{um}$, mask vectors are added to the latent sequence, allowing the autoregressive model to generate signals for masked positions.
The velocity prediction uses ODE sampling over 18 refinement steps (re-masking), taking Euler steps $\mathbf{x}^{\,i}_{t-1} = \mathbf{x}^{\,i}_t + \Delta t \cdot \mathbf{v}_{\theta}(\mathbf{x}^{\,i}_t \mid t, s^{\,i})$ of size $\Delta t$. The generator architecture is a transformer with single attention block and 16 heads, followed by the SiT-XL flow matching head. All training details are in \cref{app:training_details} and network specification is in \cref{app:nw_specs}.

\vspace{-6pt}
\subsection{Baselines, Metrics, and Evaluation Protocol}
\vspace{-2pt}
We consider recent open-sourced methods addressing the text-to-motion objective as baselines; some relevant contemporaneous approaches are omitted due to lack of public code. 
% Text2HOI \cite{cha2024text2hoi}, DiffH2O \cite{christen2024diffh2o}, LatentHOI \cite{li2024latent}, and OpenHOI \cite{zhang2025openhoi}.
Our data preprocessing mirrors the Text2HOI protocol, making their available checkpoints valid for our evaluations.
DiffH2O, assumes a grasping+interaction split, and uses the MANO PCA representation with an-object centric coordinate system (as does LatentHOI) leading to motions relative to the first frame. We therefore preprocess our data according to match their representations, but maintain our global coordinate system, and train both methods from scratch. OpenHOI inherits Text2HOI's diffusion generator, however it needs to be retrained due to mismatch in scales of supervision. 
\vspace{-6pt}

\begin{table}[t]
\centering
\scriptsize

% More horizontal space between columns
\setlength{\tabcolsep}{5.0pt}
\renewcommand{\arraystretch}{1.08}

\begin{tabular}{@{}llccccc@{}}
\toprule
Dataset
& Method
& \makecell{Acc.Top-3 $\uparrow$}
& FID $\downarrow$
& KID $\downarrow$
& Diversity $\rightarrow$
& \makecell{Matching Score $\downarrow$}
\\
\midrule

% ===================== ARCTIC =====================

\multirow{6}{*}{ARCTIC}
& \cellcolor{gray!15}\textit{Ground Truth}
& \cellcolor{gray!15}$0.991$
& \cellcolor{gray!15}--
& \cellcolor{gray!15}--
& \cellcolor{gray!15}$20.235$
& \cellcolor{gray!15}$1.295$ \\

& Text2HOI
& $0.926$ & $0.528$ & $0.008$ & $20.110$ & $2.578$ \\

& DiffH2O
& $0.715$ & $1.138$ & $0.025$ & $15.246$ & $9.791$ \\

& LatentHOI
& $0.941$ & $1.053$ & $0.009$ & $20.364$ & $3.325$ \\

& OpenHOI
& $0.946$ & $0.495$ & $0.010$ & $20.159$ & $2.292$ \\

& MAD-HOI (Ours)
& $\bm{0.991}$ & $\bm{0.071}$ & $\bm{0.004}$
& $\bm{20.312}$ & $\bm{1.353}$ \\

\cmidrule(lr){2-7}

% ===================== ARCTIC ABLATION =====================

\multirow{2}{*}{\makecell{ARCTIC\\(Ablation)}}
& w/o EOM Pred.
& $0.928$ & $1.132$ & $0.010$ & $19.467$ & $2.944$ \\

& w/o Mod. Tokens
& $0.542$ & $56.736$ & $0.787$ & $17.641$ & $11.993$ \\

\cmidrule(lr){2-7}

% ===================== ARCTIC COMPOSITE =====================

\multirow{2}{*}{\makecell{ARCTIC\\Composite}}
& \cellcolor{gray!15}\textit{Ground Truth}
& \cellcolor{gray!15}$0.890$
& \cellcolor{gray!15}--
& \cellcolor{gray!15}--
& \cellcolor{gray!15}$20.118$
& \cellcolor{gray!15}$3.320$ \\

& MAD-HOI (Ours)
& $0.869$ & $0.142$ & $0.002$ & $20.450$ & $3.693$ \\

\midrule

% ===================== GRAB =====================

\multirow{6}{*}{GRAB}
& \cellcolor{gray!15}\textit{Ground Truth}
& \cellcolor{gray!15}$0.939$
& \cellcolor{gray!15}--
& \cellcolor{gray!15}--
& \cellcolor{gray!15}$12.871$
& \cellcolor{gray!15}$3.188$ \\

& Text2HOI
& $0.745$ & $3.661$ & $0.088$ & $\bm{13.177}$ & $4.941$ \\

& DiffH2O
& $0.838$ & $1.923$ & $0.029$ & $12.250$ & $4.746$ \\

& LatentHOI
& $0.797$ & $5.616$ & $0.146$ & $13.220$ & $4.698$ \\

& OpenHOI
& $0.677$ & $4.502$ & $0.131$ & $12.316$ & $5.429$ \\

& MAD-HOI (Ours)
& $\bm{0.932}$ & $\bm{1.077}$ & $\bm{0.014}$
& $13.217$ & $\bm{4.041}$ \\

\cmidrule(lr){2-7}

% ===================== GRAB ABLATION =====================

\multirow{2}{*}{\makecell{GRAB\\(Ablation)}}
& w/o EOM Pred.
& $0.930$ & $1.471$ & $0.008$ & $14.321$ & $2.923$ \\

& w/o Mod. Tokens
& $0.737$ & $11.024$ & $0.292$ & $11.330$ & $5.772$ \\

\bottomrule
\end{tabular}

\caption{Retrieval and generation metrics on ARCTIC and GRAB.
$\downarrow$ indicates lower is better, $\uparrow$ indicates higher
is better, and $\rightarrow$ indicates closer to ground truth is better.
Best result per metric is in \textbf{bold}.}
\vspace{-6pt}
\label{tab:metrics_main}
\end{table}

\begin{table}[t]
\centering
\scriptsize
\setlength{\tabcolsep}{5pt}
\renewcommand{\arraystretch}{1.05}

\begin{adjustbox}{max width=\columnwidth}
\begin{tabular}{@{}llccccccccc@{}}
\toprule
Dataset
& Method
& Accel$\downarrow$
& Pen.\%$\downarrow$
& Con.\%$\uparrow$
& NC$\uparrow$
& IV$\downarrow$
& ID$\downarrow$
& CR$\uparrow$
& IVU$\downarrow$
& Phy$\uparrow$ \\
\midrule

% =========================================================
% ARCTIC
% =========================================================
\multirow{6}{*}{{ARCTIC}}
& \cellcolor{gray!15}\textit{Ground Truth}
& \cellcolor{gray!15}$0.18$
& \cellcolor{gray!15}$7.86$
& \cellcolor{gray!15}$80.77$
& \cellcolor{gray!15}$74.42$
& \cellcolor{gray!15}$7.41$
& \cellcolor{gray!15}$0.85$
& \cellcolor{gray!15}$13.42$
& \cellcolor{gray!15}$0.13$
& \cellcolor{gray!15}$98.67$ \\

& Text2HOI
& $0.19$
& $22.74$
& $\bm{92.90}$
& $71.77$
& $7.55$
& $0.94$
& $14.32$
& $0.15$
& $88.99$ \\

& DiffH2O
& $0.64$
& $19.98$
& $91.64$
& $73.33$
& $8.99$
& $1.03$
& $13.05$
& $0.16$
& $\bm{93.13}$ \\

& LatentHOI
& $0.31$
& $\bm{12.37}$
& $65.30$
& $57.23$
& $11.24$
& $1.15$
& $12.59$
& $0.18$
& $60.25$ \\

& OpenHOI
& $0.40$
& $20.52$
& $73.58$
& $58.48$
& $9.77$
& $1.05$
& $13.48$
& $0.17$
& $77.90$ \\

& MAD-HOI (Ours)
& $\bm{0.18}$
& \underline{$12.89$}
& \underline{$84.47$}
& $\bm{73.58}$
& $\bm{7.52}$
& $\bm{0.94}$
& $\bm{14.76}$
& $\bm{0.14}$
& \underline{$92.65$} \\

% =========================================================
% ARCTIC ABLATION
% =========================================================
\cmidrule(lr){2-11}

\multirow{2}{*}{\shortstack{{ARCTIC}\\{(Ablation)}}}
& w/o EOM Pred.
& 0.23 & 21.01 & 73.71 & 58.22 & 9.37 & 1.10 & 13.07 & 0.17 & 80.20 \\

& w/o Mod. Tokens
& 0.44 & 11.20 & 49.50 & 43.96 & 7.96 & 1.32 & 11.06 & 0.177 & 58.4 \\

% =========================================================
% ARCTIC COMPOSITE
% =========================================================
\midrule

\multirow{2}{*}{\shortstack{{ARCTIC}\\{Composite}}}
& \cellcolor{gray!15}\textit{Ground Truth}
& \cellcolor{gray!15}0.17
& \cellcolor{gray!15}7.0
& \cellcolor{gray!15}80.5
& \cellcolor{gray!15}74.86 
& \cellcolor{gray!15}8.18
& \cellcolor{gray!15}0.83
& \cellcolor{gray!15}14.38
& \cellcolor{gray!15}0.136
& \cellcolor{gray!15}98.8 \\

& MAD-HOI (Ours)
& 0.20 & 17.6 & 81.0 & 66.74 & 9.38 & 1.04 & 14.87 & 0.163 & 87.6 \\

% =========================================================
% GRAB
% =========================================================
\midrule

\multirow{6}{*}{{GRAB}}
& \cellcolor{gray!15}\textit{Ground Truth}
& \cellcolor{gray!15}$0.31$
& \cellcolor{gray!15}$30.64$
& \cellcolor{gray!15}$95.91$
& \cellcolor{gray!15}$66.52$
& \cellcolor{gray!15}$5.17$
& \cellcolor{gray!15}$2.16$
& \cellcolor{gray!15}$9.45$
& \cellcolor{gray!15}$0.14$
& \cellcolor{gray!15}$99.24$ \\

& Text2HOI
& $0.47$
& $45.11$
& $\bm{97.86}$
& $53.72$
& $9.97$
& $2.35$
& $\bm{13.59}$
& $0.19$
& $\bm{94.49}$ \\

& DiffH2O
& $0.49$
& $\bm{31.34}$
& \underline{$92.71$}
& $\bm{63.66}$
& $8.15$
& $2.51$
& $10.41$
& $0.16$
& \underline{$93.20$} \\

& LatentHOI
& $0.74$
& $35.45$
& $85.35$
& $55.09$
& $11.37$
& $2.49$
& $11.26$
& $0.21$
& $79.55$ \\

& OpenHOI
& $0.88$
& $47.11$
& $92.28$
& $48.80$
& $10.76$
& $2.50$
& $12.41$
& $0.20$
& $86.60$ \\

& MAD-HOI (Ours)
& $\bm{0.38}$
& \underline{$34.60$}
& $92.60$
& \underline{$60.56$}
& $\bm{7.98}$
& $\bm{2.07}$
& \underline{$12.42$}
& $\bm{0.15}$
& $87.01$ \\

% =========================================================
% GRAB ABLATION
% =========================================================
\cmidrule(lr){2-11}

\multirow{2}{*}{\shortstack{{GRAB}\\{(Ablation)}}}
& w/o EOM Pred.
& 0.32 & 35.6 & 90.2 & 58.08 & 7.5 & 2.14 & 11.56 & 0.17 & 83.3 \\

& w/o Mod. Tokens
& 0.59 & 24.9 & 63.0 & 47.31 & 8.45 & 2.59 & 9.15 & 0.21 & 56.4 \\

\bottomrule
\end{tabular}
\end{adjustbox}

\caption{Geometric and physical-plausibility metrics (defined in \cref{sec:metrics}). $\downarrow$ lower / $\uparrow$ higher is better; best non-GT value per column in \textbf{bold}, second-best \underline{underlined}.}
Geometric and physical-plausibility metrics (defined in
\cref{sec:metrics}). $\downarrow$ lower / $\uparrow$ higher is better;
best non-GT value per column in \textbf{bold}, second-best
\underline{underlined}.

\vspace{-12pt}
\label{tab:metrics_physics}
\end{table}

\paragraph{Metrics:} \label{sec:metrics} \textbf{\textit{For generation}}, we report Top-3 Retrieval Accuracy (R@3), FID, KID, Diversity, and Matching Score, all computed in the embedding space of our evaluator over 5 repeats; definitions follow standard practice \citep{guo2022generating} and are detailed in the appendix.
\textbf{\textit{For geometric accuracy}}, we compute Accel - mean per-vertex acceleration magnitude over the MANO mesh (scaled by 100); Penetration \% - the percentage of frames where any hand vertex penetrates the object by more than 1 cm; Contact \% - the percentage of frames with a hand vertex within 5 mm of the object surface; Net Contact - a tradeoff metric computed as $C \ast (1-P)$ to reward contact without penetration. We additionally report the fine-grained physical-plausibility suite of metrics from LatentHOI~\citep{li2024latent}, all evaluated per active hand and averaged over the frames in which that hand is in contact with the object.
{Interpenetration Volume} (IV, cm$^3$) is the volume of hand--object overlap, obtained by voxelizing the object at a 5\,mm pitch and counting voxel centers that fall inside the hand mesh (a thin-shell object convention).
{Interpenetration Depth} (ID, cm) is the deepest single hand-vertex penetration into the object.
Contact Ratio (CR, \%) is the fraction of hand vertices lying within 5\,mm of the object surface.
Interpenetration-Volume per Unit contact area (IVU, cm$^3$/cm$^2$) divides the total interpenetration volume by the total contact area, expressing penetration relative to how much of the hand actually contacts the object. And finally, Physical plausibility (Phy, \%) is the fraction of object-moving frames (mean per-vertex object displacement $>1$\,cm between consecutive frames) in which the hand is simultaneously in contact, capturing whether the hand drives object motion rather than the object drifting on its own. \textbf{\textit{For completion and infilling}}, we report Average and Final Displacement Errors ADE and FDE \citep{yuan2020dlow}. \textbf{\textit{For EOM prediction}}, the Mean and Median AE in no. of frames (with 4-frame and 8-frame tolerance variants, and and interval \%) from ground truth sequence lengths. All metric definitions are in \cref{app:metrics}.

\vspace{-8pt}
\paragraph{Evaluation Protocol:} For our generation metrics, we adopt the feature-based evaluation protocol from \citet{guo2022generating}, which learns a shared text--motion embedding space using separate text and motion encoders. The two branches are trained with a margin-based contrastive objective that pulls matched text--motion pairs together and pushes mismatched pairs apart. All prior works using this or a similar protocol adopt the full-body motion encoder for the hand trajectories only, ignoring object motion.
Since the object trajectory and articulation are key to the `action' specified in the text, we train the evaluation pipeline from scratch to encode motion latents for the hands and the object jointly, before minimizing distances between motion features and their positive text pairs. More details in \cref{app:eval_protocol}. MAD-HOI and all baselines are evaluated uniformly with this protocol.

\begin{figure*}[t]
    \centering
    \includegraphics[width=0.9\textwidth]{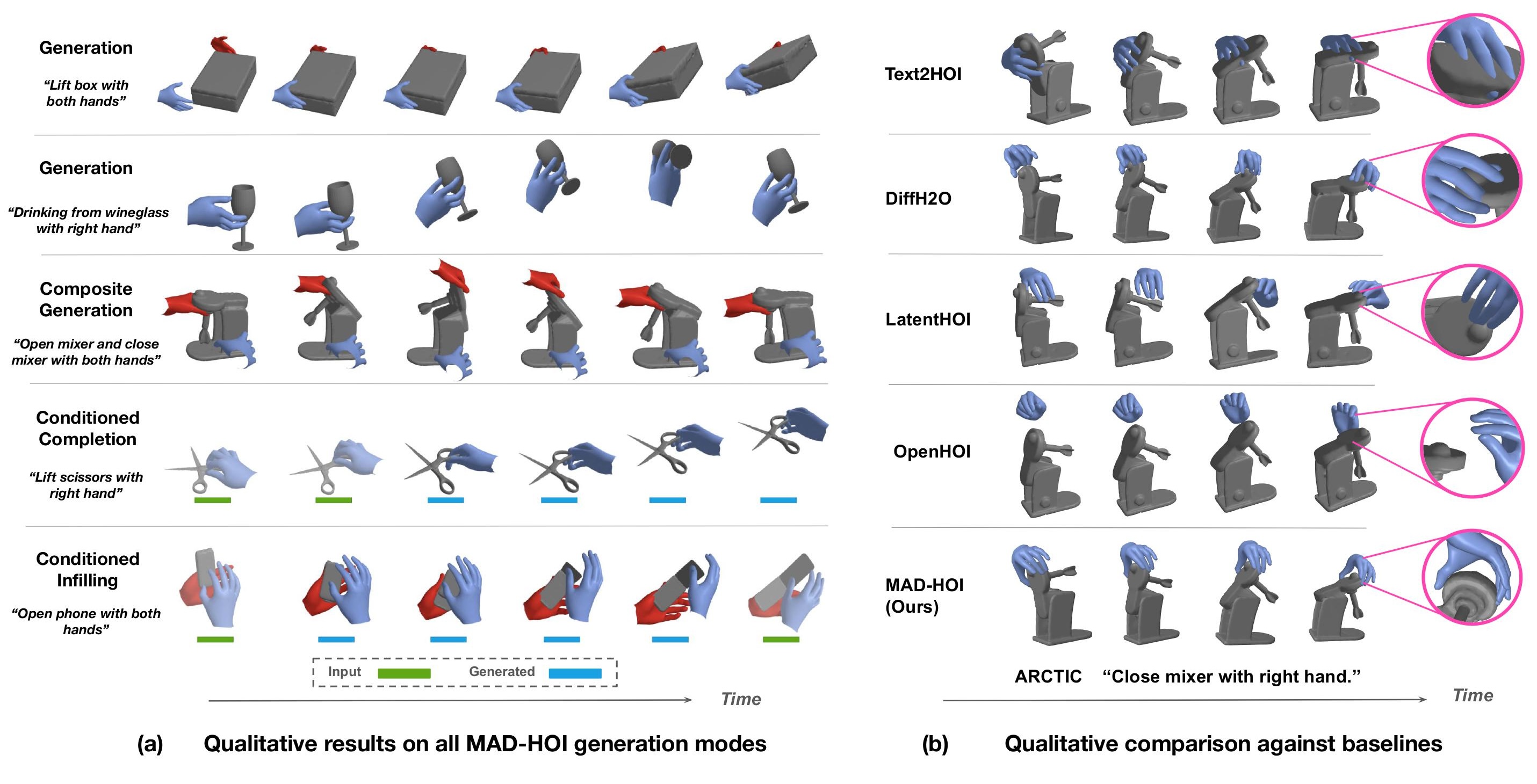}
    \vspace{-10pt}
    \caption{(a) Qualitative results showcasing all  capabilities of MAD-HOI. (b) Comparison of grasp and motion against baselines.}
    \label{fig:results_final}
    \vspace{-14pt}
\end{figure*}

\vspace{-6pt}
\section{Results}
\vspace{-8pt}
% \subsubsection{Text-to-Motion Generation}
% \vspace{-4pt}
\paragraph{Text-to-Motion Generation:} We provide qualitative results for all our capabilities and visual comparisons against baselines in \cref{fig:results_final} (GRAB comparisons in  \cref{fig:comp}). Quantitative comparisons with baselines on both metric families  are presented in \cref{tab:metrics_main} and \cref{tab:metrics_physics}.
 MAD-HOI achieves the best retrieval accuracy, FID, KID, and Matching Score on both the datasets. While it achieves the highest Diversity for ARCTIC, it trails by a small margin for GRAB, see \cref{tab:metrics_main}. OpenHOI performs the second best on ARCTIC while DiffH2O is the second best method on GRAB. MAD-HOI's generations show the most gains in text-alignment metrics like FID and Matching Score.  %\textcolor{red}{segregate results into primary and secondary metrics to avoid questions on ablations missing in table 2}

When it comes to geometric metrics (\cref{tab:metrics_physics}), we note that MAD-HOI performs well on ARCTIC. It has the best performance on 6 out of the 9 metrics, coming in second on Physical Plausibility, Penetration and Contact $\%$s. We realize that since contact and penetration exist in a tradeoff, we compute NC to show a balance between the two and here MAD-HOI has the highest NC value. The picture is more blurry for GRAB though. MAD-HOI still has the smoothest generated motions (with the lowest acceleration and closer to ground truth (GT) as with ARCTIC). However, it trails in Contact $\%$ and CR. We still outperform the baselines in IV, ID and IVU and clock in the second best values for Penetration $\%$, NC. However, the lower physical plausibility points to loss of contact when the object is off the ground plane and moving fast -  something more characteristic of the GRAB dataset where the object trajectories span large areas. Overall, MAD-HOI is the strongest method for text-aligned generation with competitive geometric metrics. No baseline method outperforms MAD-HOI across both metric families consistently. It is also the preferred generation method in our user study discussed in \cref{app:user_study}.

\vspace{-8pt}
\paragraph{Composite Motion Generation:} The MAD-HOI generator is trained with bidirectional attention and is pushed to predict EOM tokens. This coupled with  iterative masked denoising at inference, gives MAD-HOI the ability to generate motions of varying lengths within its max context window, meaning that it can generate composite motions based on chained prompts up to 152 frames. We train the MAD-HOI generator on a bigger ARCTIC training dataset where we include 30\% of the 2197 compound motions (up to 4 atomic motions chained). This debiases EOM likelihoods from atomic motion lengths and introduces words like 'and' to the generator. Many compound combinations are unseen in validation and metrics from \cref{tab:metrics_main} and \cref{tab:metrics_physics} show that MAD-HOI has minimal quality degradation in composite generations. We omit this experiment on GRAB: its sequences are longer leaving very few valid testing sequences below the maximum context window.

\vspace{-8pt}
\paragraph{Conditioned Completion and Infilling:} During inference, masked out tokens beyond a few starting frames in the sequence result in completion, and in the center, result in infilling.
Our generator is context-aware, its latent prediction is conditioned on whatever unmasked input is provided. We therefore report the ADE and FDE for masked frames with the closest semantic GT samples rather than raw trajectory errors in meters.
In (\cref{tab:DE}), errors largely decrease monotonically as more ground-truth latents are provided to the generator as context, as is expected. At 2 conditioning latents, completion quality is close to atomic generation while infilling is harder, due to the stricter constraint of matching context on both sides. Our HOI baselines do not perform completion/infilling, we therefore include simple protocol-matched kinematic baselines in \cref{app:comp_inf_baseline}.

\vspace{-8pt}
\paragraph{Sequence Termination with EOM:}
When the MAD-HOI generator predicts latents close to the learned EOM token for the masked tokens of all 3 modalities, the end of the sequence is detected and the trailing frames are discarded as the sequence is terminated. No separate length predictor needs to be trained or queried.
We compare our implicit sequence termination performance indirectly by comparing against the dedicated cVAE length regressor of Text2HOI in \cref{tab:eom_metrics1} which is explicitly trained with supervision on sequence lengths.
The specialist regressor is more accurate on raw error, as expected. A single prompt, however, maps to many valid durations, so we also evaluate against the $[q_{10}, q_{90}]$ interval of demonstrated lengths per condition: MAD-HOI's predicted endpoints fall inside this valid range for 78.4\% (ARCTIC) and 70.5\% (GRAB) of prompts, and miss it by only 2--5 frames on average.
Termination thus comes at no extra cost to the single generation objective or number of parameters, with modest accuracy against a specialist module. \cref{app:learned_eom} outlines why learned EOM is better than user-specified EOM.

% \begin{table}[H]
% \centering

% \small
% \setlength{\tabcolsep}{3pt}

% \begin{tabular}{
% @{}ll
% *{5}{c}
% @{\hspace{8pt}}
% *{5}{c}
% @{\hspace{8pt}}
% c
% @{\hspace{8pt}}
% c@{}
% }
% \toprule
% & &
% \multicolumn{5}{c}{ADE $\downarrow$}
% &
% \multicolumn{5}{c}{FDE $\downarrow$}
% &
% FID $\downarrow$
% &
% Diversity $\uparrow$
% \\

% \cmidrule(lr){3-7}
% \cmidrule(lr){8-12}
% \cmidrule(lr){13-13}
% \cmidrule(lr){14-14}

% \multicolumn{2}{@{}l}{\textbf{Task Tokens (\#):}}
% & 1 & 2 & 3 & 4 & 5
% & 1 & 2 & 3 & 4 & 5
% & 2
% & 2
% \\
% \midrule

% \multirow{2}{*}{Completion}
% & ARCTIC
% & 0.638 & 0.611 & 0.618 & 0.608 & 0.605
% & 0.811 & 0.786 & 0.796 & 0.737 & 0.742
% & 0.173
% & 20.325 
% \\

% & GRAB
% & 0.791 & 0.782 & 0.759 & 0.765 & 0.747
% & 0.995 & 0.965 & 1.018 & 0.931 & 0.863
% & 1.719
% & 13.010
% \\
% \midrule

% \multirow{2}{*}{Infilling}
% & ARCTIC
% & 0.747 & 0.678 & 0.646 & 0.619 & 0.612
% & 0.792 & 0.739 & 0.700 & 0.671 & 0.664
% & 0.182
% & 19.2136
% \\

% & GRAB
% & 0.739 & 0.586 & 0.567 & 0.536 & 0.505
% & 0.870 & 0.655 & 0.615 & 0.607 & 0.547
% & 1.931
% & 13.649
% \\
% \bottomrule
% \end{tabular}
% \caption{Evaluation of conditioned completion and infilling.
% Lower ADE, FDE, and FID are better,  higher Diversity is better.}
% \label{tab:DE}
% \vspace{-10pt}
% \end{table}

\begin{table}[H]
\centering

\scriptsize
\setlength{\tabcolsep}{2.2pt}
\renewcommand{\arraystretch}{1.05}

\begin{adjustbox}{max width=\columnwidth}
\begin{tabular}{
@{}ll
@{\hspace{12pt}}     % after dataset column
*{5}{c}
@{\hspace{14pt}}     % ADE -> FDE
*{5}{c}
@{\hspace{12pt}}     % FDE -> FID
c
@{\hspace{10pt}}     % FID -> Diversity
c@{}
}
\toprule

& &
\multicolumn{5}{c}{ADE $\downarrow$}
&
\multicolumn{5}{c}{FDE $\downarrow$}
&
FID $\downarrow$
&
Diversity $\uparrow$
\\

\cmidrule(lr){3-7}
\cmidrule(lr){8-12}
\cmidrule(lr){13-13}
\cmidrule(lr){14-14}

\multicolumn{2}{@{}l}{\textbf{Task Tokens (\#):}}
& 1 & 2 & 3 & 4 & 5
& 1 & 2 & 3 & 4 & 5
& 2
& 2
\\

\midrule

\multirow{2}{*}{Completion}
& ARCTIC
& 0.649 & 0.617 & 0.613 & 0.608 & 0.592
& 0.816 & 0.787 & 0.789 & 0.737 & 0.733
& 0.173
& 20.325
\\

& GRAB
& 0.748 & 0.776 & 0.738 & 0.749 & 0.724
& 0.963 & 1.019 & 1.001 & 0.960 & 0.882
& 1.719
& 13.010
\\

\midrule

\multirow{2}{*}{Infilling}
& ARCTIC
& 0.783 & 0.707 & 0.675 & 0.639 & 0.629
& 0.833 & 0.766 & 0.719 & 0.691 & 0.684
& 0.182
& 19.2136
\\

& GRAB
& 0.721 & 0.579 & 0.546 & 0.514 & 0.484
& 0.844 & 0.671 & 0.614 & 0.591 & 0.536
& 1.931
& 13.649
\\

\bottomrule
\end{tabular}
\end{adjustbox}
\vspace{-4pt}
\caption{
Evaluation of conditioned completion and infilling.
Lower ADE, FDE, and FID are better; higher Diversity is better.
}
\label{tab:DE}
\vspace{-12pt}
\end{table}

% \begin{table}[H]
% \centering
% \scriptsize
% \setlength{\tabcolsep}{1.2pt}
% \renewcommand{\arraystretch}{1.05}

% \begin{adjustbox}{max width=\columnwidth}
% \begin{tabular}{@{}lcccc@{}}
% \toprule
% & \multicolumn{2}{c}{\textbf{ARCTIC}}
% & \multicolumn{2}{c}{\textbf{GRAB}} \\
% \cmidrule(lr){2-3}
% \cmidrule(lr){4-5}

% \textbf{Metric}
% & \shortstack{Text2HOI\\cVAE}
% & \shortstack{MAD-HOI\\(Ours)}
% & \shortstack{Text2HOI\\cVAE}
% & \shortstack{MAD-HOI\\(Ours)} \\
% \midrule

% MAE $\downarrow$
% & 7.92 & 12.23 & 14.75 & 20.43 \\

% MedAE $\downarrow$
% & 5.00 & 8.00 & 11.20 & 14.40 \\

% Within-4 (\%) $\uparrow$
% & 44.2 & 44.3 & 31.1 & 33.1 \\

% Within-8 (\%) $\uparrow$
% & 67.2 & 62.8 & 43.0 & 38.3 \\

% Interval MAE $\downarrow$
% & 0.26 & 2.32 & 1.02 & 4.85 \\

% Within Interval (\%) $\uparrow$
% & 90.7 & 78.4 & 80.0 & 70.5 \\

% \bottomrule
% \end{tabular}
% \end{adjustbox}
% \caption{Quantitative evaluation of EOM prediction on ARCTIC
% and GRAB. We report regression errors, threshold-based accuracies, and
% distribution-aware tolerance metrics.}
% \label{tab:eom_metrics1}
% \vspace{-10pt}
% \end{table}

\begin{table}[H]
\centering
\begin{minipage}[c]{0.48\linewidth}
\centering
\scriptsize
\setlength{\tabcolsep}{1.2pt}
\renewcommand{\arraystretch}{1.05}
\begin{adjustbox}{max width=\linewidth}
\begin{tabular}{@{}lcccc@{}}
\toprule
& \multicolumn{2}{c}{\textbf{ARCTIC}}
& \multicolumn{2}{c}{\textbf{GRAB}} \\
\cmidrule(lr){2-3}
\cmidrule(lr){4-5}
\textbf{Metric}
& \shortstack{Text2HOI\\cVAE}
& \shortstack{MAD-HOI\\(Ours)}
& \shortstack{Text2HOI\\cVAE}
& \shortstack{MAD-HOI\\(Ours)} \\
\midrule
MAE $\downarrow$             & 7.92 & 12.23 & 14.75 & 20.43 \\
MedAE $\downarrow$           & 5.00 & 8.00  & 11.20 & 14.40 \\
Within-4 (\%) $\uparrow$     & 44.2 & 44.3  & 31.1  & 33.1  \\
Within-8 (\%) $\uparrow$     & 67.2 & 62.8  & 43.0  & 38.3  \\
Interval MAE $\downarrow$    & 0.26 & 2.32  & 1.02  & 4.85  \\
Within Interval (\%) $\uparrow$ & 90.7 & 78.4 & 80.0 & 70.5 \\
\bottomrule
\end{tabular}
\end{adjustbox}
\caption{EOM prediction on ARCTIC and GRAB: regression errors,
threshold accuracies, and distribution-aware tolerance metrics.}
\label{tab:eom_metrics1}
\end{minipage}\hfill
\begin{minipage}[c]{0.49\linewidth}
\centering
\includegraphics[width=\linewidth,height=3.0cm,keepaspectratio]{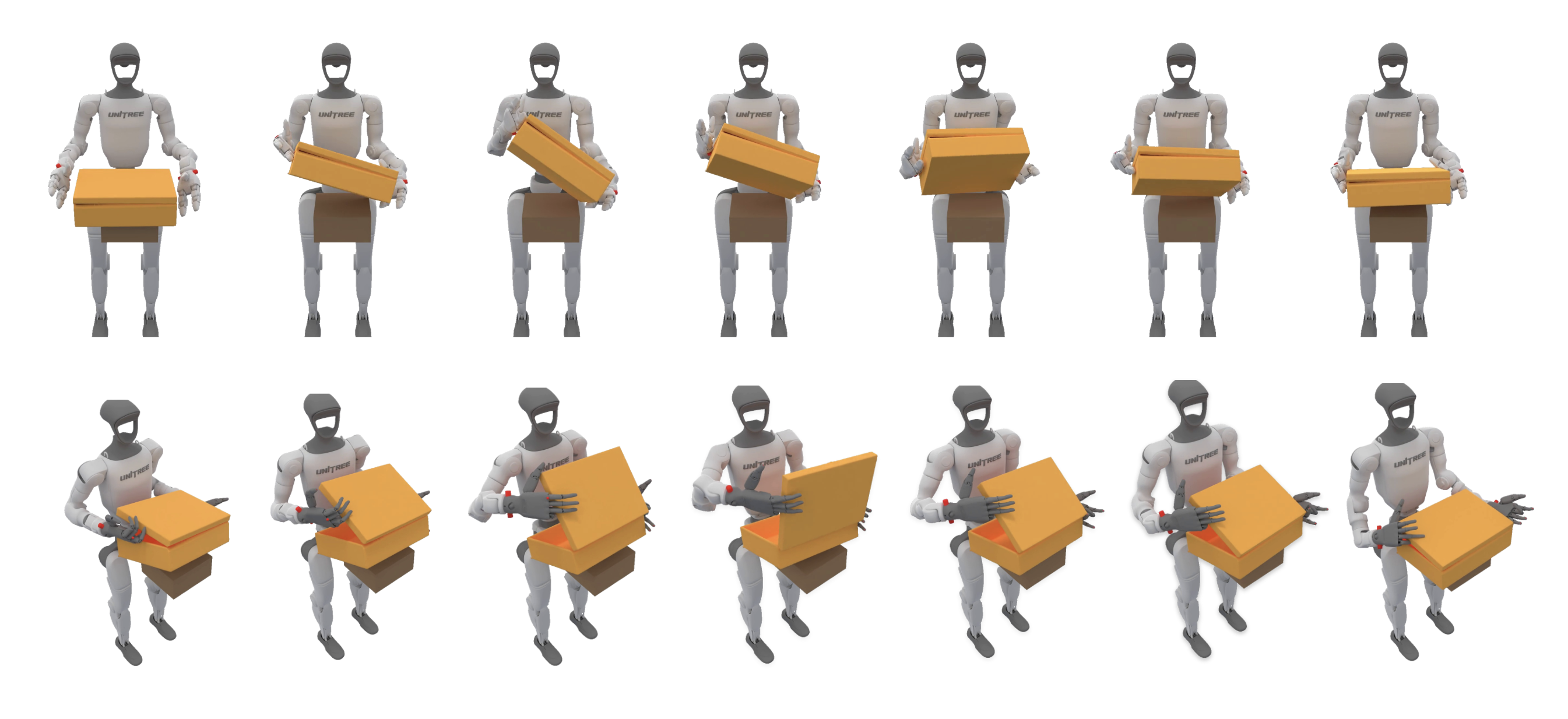}
\vspace{-4pt}
\figcaption{MAD-HOI output retargeted to a Unitree G1 with Inspire hands
(\cref{app:application_robot}).}
\label{fig:applications_inspire}
\end{minipage}
\vspace{-12pt}
\end{table}

\paragraph{Ablation Experiments:}
We ablate the two design decisions our capability claims rest on: EOM prediction and the learnable modality embeddings that disentangle the three streams.
We switch off each individually in our base model and test for atomic generations. As seen in (\cref{tab:metrics_main}), modality embeddings are necessary for convergence: without them, FID degrades from $0.071$ to $56.7$ on ARCTIC and from $1.125$ to $11.02$ on GRAB, and the network cannot learn a usable distribution. Switching off EOM prediction is comparatively mild but still costly: FID rises from $0.071$ to $1.13$ on ARCTIC and from $1.125$ to $1.47$ on GRAB. Geometric metrics also echo similar trends for both datasets. Thus, learning explicit termination does not burden generation and appears to help the model organize varying lengths for similar prompts. The appendix also provides ablations on the geometric losses used for VAE training \cref{tab:loss_ablation}. 
We further present results from a generalization experiment with MAD-HOI in \cref{app:general}.

\vspace{-8pt}
\section{Conclusion}
\vspace{-6pt}
We present MAD-HOI, a masked autoregressive generation framework that produces articulated HOI sequences from text with true  handedness. It is further able to generate composite motions, complete and infill motions with minimal degradation in quality. The unique structure of the continuous latent space from our VAE leads to stream-wise modality control and enables learning reliable end-of-motion, which in turn improves generation. In ARCTIC and GRAB, MAD-HOI outperforms retrained open-source baselines on distributional fidelity while being competitive on geometric accuracy and physical plausibility. Finally, its outputs can reliably seed functional retargeting pipelines for humanoid robots (\cref{fig:applications_inspire}; details in \cref{app:application_robot}).

\bibliography{iclr2027_conference}
\bibliographystyle{iclr2027_conference}
\clearpage
\appendix
\begin{bibunit}
% \crefname{section}{Appendix}{Appendices}
% \Crefname{section}{Appendix}{Appendices}
% \crefname{subsection}{Appendix}{Appendices}
% \Crefname{subsection}{Appendix}{Appendices}
% \crefname{subsubsection}{Appendix}{Appendices}
% \Crefname{subsubsection}{Appendix}{Appendices}

\section{Appendix}
In this appendix, we provide a review of literature in the larger field of diffusion, jerk profiles for motion reconstructions from discrete and continuous latents, followed by technical details about our method and training (loss definitions, the input token sequence example, training details, and network specifications). We then present details of our evaluation pipeline and metrics, qualitative samples from GRAB, followed by experiments to further validate our claims. Finally, we show our method being applied to a humanoid robot and discuss future work

% \textbf{In the supplementary material, we present interactive 3D visualizations for all our generation capabilities and text to motion from competing baselines. View html file named 'visualize\_hoi\_motions' under the provided zipped folder. OpenHOI omitted for space constraints and for similar performance to its stem - Text2HOI.  }

\vspace{-4pt}
\subsection{Related Works:}
\textbf{Full-body motion generation.}
Diffusion-based methods such as MDM~\citep{tevet2022human}, MotionDiffuse~\citep{zhang2024motiondiffuse}, and MLD~\citep{chen2023executing} model continuous action- or text-conditioned trajectories and generate smooth, diverse motions. In parallel, TM2T~\citep{guo2022tm2t}, T2M-GPT~\citep{zhang2023t2m}, MotionGPT~\citep{jiang2023motiongpt}, and MoMask~\citep{guo2024CVPR} represent motion using discrete VQ tokens, enabling flexible sequence modeling at the cost of quantization. PriorMDM~\citep{shafir2024human} further demonstrates diffusion-based long-duration and spatio-temporal composition, while MARDM~\citep{meng2025rethinking} and MoMADiff~\citep{zhang2025towards} combine continuous diffusion with masked autoregressive generation. 

\vspace{-6pt}
\paragraph{Evolution of diffusion-based sequence modeling.}
Diffusion models generate continuous data by reversing a progressive
noising process~\citep{sohl2015deep,ho2020denoising}. Flow-matching and
stochastic-interpolant formulations instead learn a continuous transport
field between a simple noise distribution and the data distribution;
SiT~\citep{ma2024Sit}, for example, parameterizes this velocity field with
a Transformer architecture. Recent work has combined these continuous generative objectives with
sequence modeling. Diffusion Forcing~\citep{chen2024diffusion} assigns
independent noise levels to causally ordered sequence elements, allowing
past tokens to provide cleaner context while future tokens remain more
strongly corrupted. MAR~\citep{li2024autoregressive}, by contrast, extends
masked autoregressive modeling to continuous representations: a
Transformer predicts contextual features for masked positions, while a
diffusion decoder generates the corresponding continuous tokens.

These ideas have subsequently been adapted to motion synthesis.
MARDM~\citep{meng2025rethinking} combines bidirectional masked
autoregression with a per-token diffusion sampler, while
MoMADiff~\citep{zhang2025momadiff} employs masked autoregressive diffusion
for controllable text-to-motion generation. MAD-HOI builds on this
continuous masked-autoregressive paradigm, but structures the sequence
into separate object, left-hand, and right-hand latent streams and uses
a flow-matching head to generate masked motion latents. This formulation
enables generation, completion, infilling, modality-specific control,
and learned termination within a single model.

\vspace{-4pt}
\subsection{Tokenization: Discrete vs Continuous}
\begin{figure*}[t]
\centering
\textbf{}\par
\vspace{0.3em}
\includegraphics[width=\linewidth]{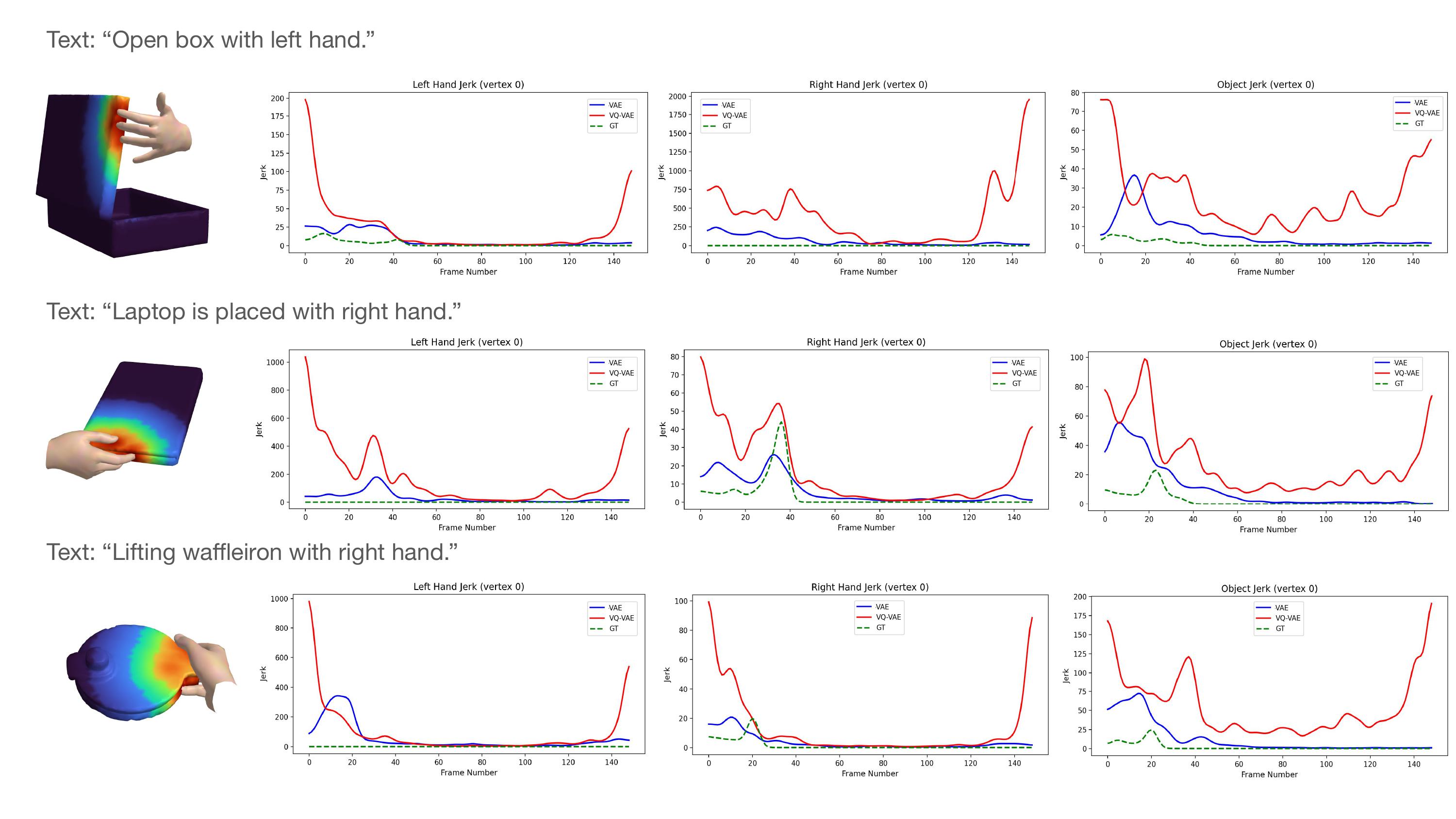}
\caption{VAE vs VQ-VAE Reconstruction Jitter for ARCTIC: Jerk Analysis for Smoothness Comparison. We plot jerk magnitude for ground truth, VAE-reconstructed, and VQ-VAE-reconstructed trajectories of the MANO hand wrist joints and object centers. }
\label{img:jitter1}
\end{figure*}

\begin{figure*}[t]
\centering
\textbf{}\par
\vspace{0.3em}
\includegraphics[width=\linewidth]{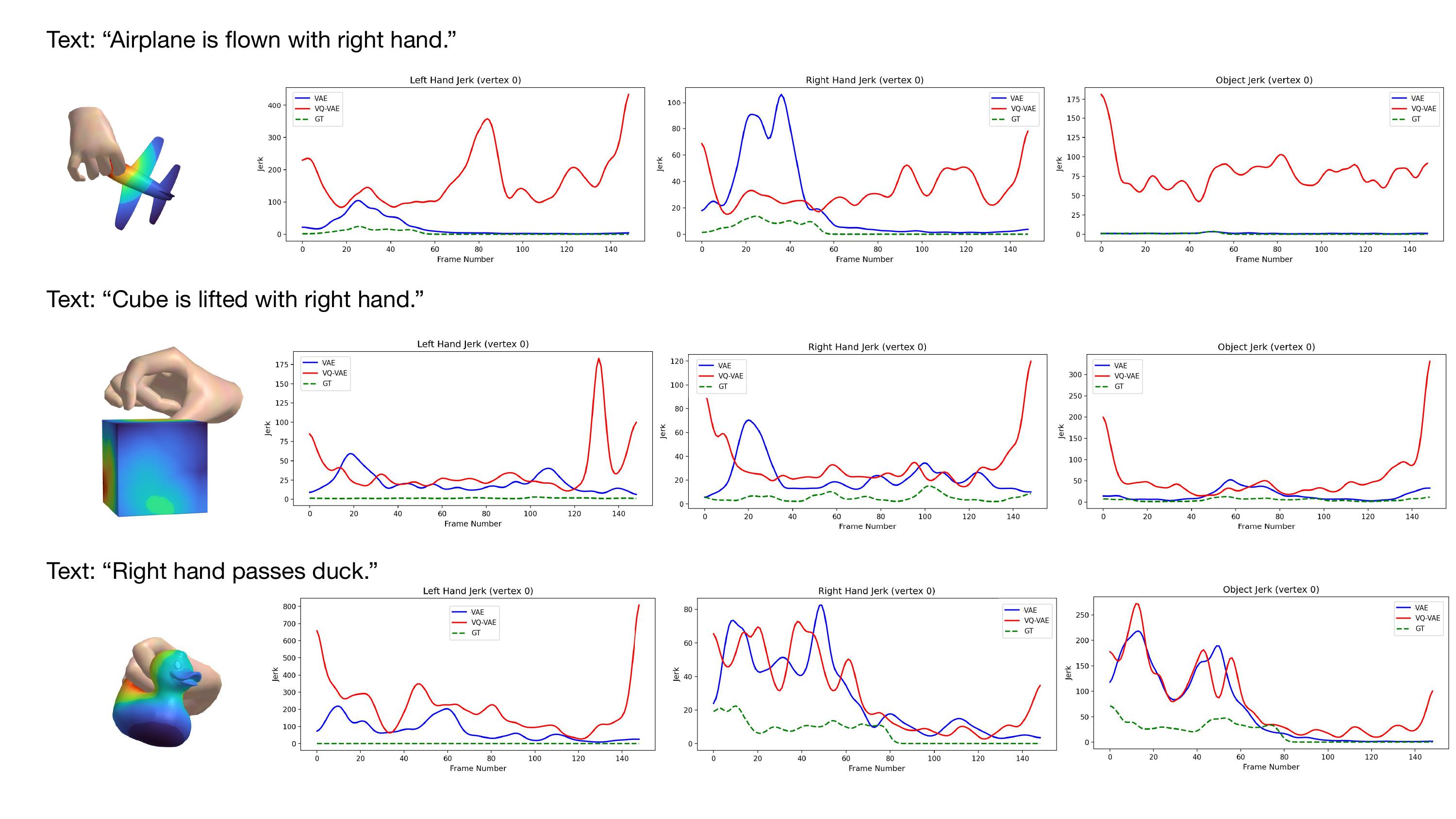}
\caption{VAE vs VQ-VAE Reconstruction Jitter for GRAB: Jerk Analysis for Smoothness Comparison. We plot jerk magnitude for ground truth, VAE-reconstructed, and VQ-VAE-reconstructed trajectories of the MANO hand wrist joints and object centers.}
\label{img:jitter2}
\end{figure*}

\label[appendix]{app:disc_v_cont}
\vspace{-4pt}
We argue that motions reconstructed from discrete codes lose fine-grained information and this directly results in motions that are less smooth and more jittery. We take inspiration from \citep{Cho_2025_ICCV}, and assess the temporal coherence of the reconstructed motions by analyzing their jerk profiles, i.e., the third derivative of position over time. Jerk quantifies fine-grained temporal irregularities, allowing us to distinguish overly static motion from high-frequency jitter. Lower and smoother jerk magnitudes correspond to more natural and physically consistent trajectories. See \cref{img:jitter1} and \cref{img:jitter2} for plots on jerk profiles of ground truth and reconstructed trajectories of the wrist joints and the object centers obtained from the VAE and its equivalent VQ-VAE for samples from ARCTIC and GRAB respectively. We observe that the VQ-VAE-reconstructed trajectories are less smooth, with spikes corresponding to abrupt, non-physical changes. The trajectories from the VAE are closer in smoothness to the ground truth data.

\subsection{Method Details}

\subsubsection{Losses for VAE Training } \label{sub:loss_defs}

Our VAE is trained to reconstruct fine-grained motions and hand-object contacts. To facilitate this, we apply some geometric losses that encourage close contact and penalize penetration. They are defined as follows:

\textbf{Contact Loss:} We enforce a threshold-based contact loss. $D(J_i^o)$ calculates the distance between a hand joint and the closest object surface, enforcing the threshold distance constraint $D(J_i^o) \le \phi$:
\begin{equation}
\mathcal{L}_C = \sum_i D(\hat{J}_i^o), \quad \forall\, D(\hat{J}_i^o) \le \phi,
\label{eq:contact_loss}
\end{equation}
where
\begin{equation}
D(\hat{J}_i^o) = \min_j \left\lVert V_j^p - \hat{J}_i^o \right\rVert_2^2,
\end{equation}
with $V_j^p$ representing object surface points and $J_i^o$ representing hand joints.
\vspace{4pt}

\textbf{Penetration Loss:} To penalize penetration of the hand mesh and the object mesh, we minimize the average squared distances of penetrating hand vertices $p \in P_{\text{in}}^{o}$ to their closest object vertex $\hat{V}_i^{\,o}$:
\begin{equation}
\mathcal{L}_{\text{pen}}
= \frac{1}{\lvert P_{\text{in}}^{o} \rvert}
\sum_{p \in P_{\text{in}}^{o}}
\min_i \, \lVert p - \hat{V}_i^{\,o} \rVert_2^{2}.
\label{eq:pen_loss}
\end{equation}

\textbf{Distance Map Loss:} Following \citep{liang2024intergen}, we apply a distance map loss to align the estimated distance map (distance from each hand joint to the closest object vertex) with the ground-truth distance map. For a single hand,
\begin{equation}
\mathcal{L}_{\mathrm{DM}}
= \sum_{i} \bigl(D(\hat{J}^o_i) - D({J}^o_i)\bigr)^2 \,\mathbf{1}\bigl[D({J}^o_i) < \phi\bigr],
\label{eq:ldm_single_hand}
\end{equation}
where $\phi$ is a threshold that selects near-contact regions.

We also study the effect of these losses on the reconstruction quality of our VAE. See \cref{tab:loss_ablation} to see the ablated performance of our VAE on the GRAB dataset. Without the penetration loss, there is high contact but it it assuredly comes with penetration. Without the contact loss, the contact \% is lower than optimum. Perhaps the most important loss is the Distance Map Loss, as when its removed the contact worsens severely, and further the reconstruction losses and MPJPE is also impacted. The combination of all losses performs the best. 

\begin{table}[t]
\centering
\small
\setlength{\tabcolsep}{6pt}
\begin{tabular}{l|cccc}
\toprule
\rowcolor{gray!15}
\textbf{Metric}
& w/o $\mathcal{L}_{\mathrm{Pen}}$
& w/o $\mathcal{L}_{\mathrm{C}}$
& w/o $\mathcal{L}_{\mathrm{DM}}$
& \textbf{All losses} \\
\midrule
Obj.\ L1 $\downarrow$   & 0.005 & 0.010 & 0.018 & \textbf{0.005} \\
Hand L1 $\downarrow$    & 0.013 & 0.023 & 0.076 & \textbf{0.013} \\
MPJPE $\downarrow$      & 0.049 & 0.052 & 0.119 & \textbf{0.044} \\
Pen.\ (\%) $\downarrow$ & 61.16 & 34.94 & 0.00  & \textbf{2.10} \\
Con.\ (\%) $\uparrow$   & 98.83 & 79.48 & 22.4  & \textbf{92.20} \\
\bottomrule
\end{tabular}
\caption{\textbf{VAE Training on GRAB}: Ablation of the geometric loss terms .}
\label{tab:loss_ablation}
\end{table}

\subsubsection{Input Token Sequence to Transformer Example} \label[appendix]{app:seq}
If we assume a max token length of 18 (6 possible latent timesteps x 3 for modalities), and consider a motion having 5 valid latent timesteps, where we mask one latent timestep out and the EOM tokens are appended at the end, the sample input token sequence to the transformer would look like \cref{eq:inp_tok}.
% {
% \small

% \begin{equation}
% \begin{gathered}
% \langle z_o^0 \rangle\ \langle z_l^0 \rangle\ \langle z_r^0 \rangle\ 
% \langle z_o^1 \rangle\ \langle z_l^1 \rangle\ \langle z_r^1 \rangle\ 
% \langle \text{MASK} \rangle\ \langle \text{MASK} \rangle\ \langle \text{MASK} \rangle 
% \langle z_o^3 \rangle\ \langle z_l^3 \rangle\ \langle z_r^3 \rangle\ 
% \langle z_o^4 \rangle\ \langle z_l^4 \rangle\ \langle z_r^4 \rangle\
% \langle \text{EOM}_o \rangle\ \langle \text{EOM}_l \rangle\ \langle \text{EOM}_r \rangle
% \end{gathered}
% \label{eq:inp_tok}
% \end{equation}}

\begin{equation}
\resizebox{0.96\columnwidth}{!}{$
\displaystyle
\langle z_o^0 \rangle\,
\langle z_l^0 \rangle\,
\langle z_r^0 \rangle\,
\langle z_o^1 \rangle\,
\langle z_l^1 \rangle\,
\langle z_r^1 \rangle\,
\langle \text{MASK} \rangle\,
\langle \text{MASK} \rangle\,
\langle \text{MASK} \rangle\,
\langle z_o^3 \rangle\,
\langle z_l^3 \rangle\,
\langle z_r^3 \rangle\,
\langle z_o^4 \rangle\,
\langle z_l^4 \rangle\,
\langle z_r^4 \rangle\,
\langle \text{EOM}_o \rangle\,
\langle \text{EOM}_l \rangle\,
\langle \text{EOM}_r \rangle
$}
\label{eq:inp_tok}
\end{equation}

\subsubsection{Flow-Matching Details}
\label[appendix]{app:fm_details}
Given the linear interpolant $x_t = \alpha_t x_0 + \sigma_t \epsilon$ from the main paper, with $\alpha_t = 1-t$ and $\sigma_t = t$, the velocity field toward which the diffusion MLP regresses is defined as
\begin{equation}
v(x, t) = \dot{\alpha}_t\, \mathbb{E}[x_0 \mid x_t = x] + \dot{\sigma}_t\, \mathbb{E}[\epsilon \mid x_t = x],
\label{eq:velocity}
\end{equation}
where $\dot{\alpha}_t$ and $\dot{\sigma}_t$ are the derivatives of the interpolation coefficients with respect to $t$.

\subsubsection{Training Details} \label[appendix]{app:training_details}

We train the generation model on 4 H100 GPUs for 200000 iterations for ARCTIC and 150000 iterations for GRAB with a max learning rate of 3e-4, which is achieved through a linear warmup over 1000 iterations.
The learning rate then decays with cosine annealing to 1e-4.
The VAE is trained on 2 H100 GPUs for 6000 epochs with a learning rate of 2e-4, with $\alpha=1$, $\beta=0.5$, $\gamma=0.5$, $\delta=1$. The KL term is internally scaled by
$\beta_{\mathrm{KL}}=10^{-4}$ and subsequently weighted by
$\lambda_{\mathrm{KL}}=0.5$. We use this small coefficient because the KL divergence is summed over the latent-channel dimension before being averaged over valid temporal tokens.
Our VAE operates over short temporal sequences, as opposed to static poses. We use a window length of 152 to maximize the temporal context within the VAE and have a 4 frames to 1 token (latent) ratio throughout the method.

The longer window length presents some challenges when dealing with a dataset like ARCTIC, which has many shorter atomic motion sequences.
For shorter sequences, we pad the last valid frame until the end and also maintain replicate padding within the VAE to ensure the network does not see zeros or abrupt signal changes, which could lead to drift-like artifacts during decoding.
Further, the VAE reconstruction losses are computed over the padded sequences as opposed to the valid-only frames, forcing the VAE to learn to reconstruct static hands and objects from idle latents.
This design complements the EOM mechanism (see main paper \cref{subsec:ard}).

\begin{table}[t]
  \centering
  \small
  \setlength{\tabcolsep}{3.5pt}
  \renewcommand{\arraystretch}{1.05}
  
  \begin{tabularx}{\columnwidth}{@{}l>{\raggedright\arraybackslash}Xl@{}}
  \toprule
  \textbf{Layer} & \textbf{Operation} & \textbf{Output} \\
  \midrule
  Input & Object sequence & $B \times T \times C_{obj}$ \\
  Preprocess & permute$(B,T,C_{obj}) \rightarrow (B,C_{obj},T)$
  & $B \times C_{obj} \times T$ \\
  Encoder & Conv1d$(C_{obj} \rightarrow D, k{=}3, s{=}1, p{=}1)$
  + ReLU & $B \times D \times T$ \\
  Encoder block & Conv1d$(D \rightarrow D, k{=}4, s{=}2, p{=}1)$
  + ResNet1D & $B \times D \times T/2$ \\
  Encoder block & Conv1d$(D \rightarrow D, k{=}4, s{=}2, p{=}1)$
  + ResNet1D & $B \times D \times T_{lat}$ \\
  Latent proj. & Conv1d$(D \rightarrow D, k{=}3, s{=}1, p{=}1)$
  & $B \times D \times T_{lat}$ \\
  Posterior mean & Conv1d$(D \rightarrow D, k{=}1)$ & $\mu:\ B
  \times D \times T_{lat}$ \\
  Posterior log-var & Conv1d$(D \rightarrow D, k{=}1)$, clamp
  $[-10,10]$ & $\log\sigma^2:\ B \times D \times T_{lat}$ \\
  Sampling & $z = \mu + \epsilon \odot
  \exp(\tfrac{1}{2}\log\sigma^2)$ & $B \times D \times T_{lat}$
  \\
  Concat & concat$(z, f_{pointnet})$ & $B \times (D{+}P) \times
  T_{lat}$ \\
  Decoder & Conv1d$(D{+}P \rightarrow D)$ + ReLU & $B \times D
  \times T_{lat}$ \\
  Decoder block & ResNet1D $\rightarrow$ Upsample $\rightarrow$
  Conv1d & $B \times D \times T/2$ \\
  Decoder block & ResNet1D $\rightarrow$ Upsample $\rightarrow$
  Conv1d & $B \times D \times T$ \\
  Decoder & Conv1d$(D \rightarrow D)$ + ReLU & $B \times D
  \times T$ \\
  Output & Conv1d$(D \rightarrow C_{obj})$ & $B \times C_{obj}
  \times T$ \\
  Postprocess & permute$(B,C_{obj},T) \rightarrow (B,T,C_{obj})$
  & $B \times T \times C_{obj}$ \\
  \bottomrule
  \end{tabularx}
  \caption{Network architecture of the object stream of the
  VAE.}
  \label{tab:objvae_arch}
  \end{table}

\subsection{Network Architecture Specifications}
\label[appendix]{app:nw_specs}
\subsubsection{VAE}
\noindent Shape symbols: $B$: batch size, $T$: input number of frames, \texttt{$T_{lat}$}: number of tokens where \texttt{$T_{lat} = T / 2^{down\_t} $} and \texttt{down\_t}=2, $C\_obj$: channels in object feature (10 for ARCTIC and 9 for GRAB), $C\_hand$: channels in hand feature (100), $D$: latent dimension (512), $P$: channels in pointnet features (1024), $mu$: distribution mean and $\sigma$: log variance. See \cref{tab:objvae_arch} and \cref{tab:handae_arch} for the specification tables.

\subsubsection{Generator}
For our generator, we use a single attention block with 16 heads and follow it up with an SiT-XL network which has 16 residual blocks. The flow matching head is queried for 18 refinement steps where random tokens are denoised repeatedly.

Shape symbols: $B$: batch size, $L$: sequence length (number of max motion tokens = 117; 38 tokens x 3 modalities + 3 EOM tokens), $D_{ae}$: AE latent dimension (512), $D$: transformer hidden dimension (1024),
$D_{ff}$: transformer feed-forward hidden dimension (4096),
$H$: number of attention heads (16),
$N_{tr}$: number of MAR transformer blocks (1),
$N$: number of masked/selected tokens processed by DiffMLPs (variable), $W_{diff}$: DiffMLPs hidden width (1792),
$N_{diff}$: number of DiffMLPs residual blocks (16). See \cref{tab:gen_arch} for the specification table.

\begin{table}[t]
\centering
\small
\setlength{\tabcolsep}{3.5pt}
\renewcommand{\arraystretch}{1.05}

\begin{tabularx}{\columnwidth}{@{}l>{\raggedright\arraybackslash}Xl@{}}
  \toprule
  \textbf{Layer} & \textbf{Operation} & \textbf{Output} \\
  \midrule
  Input & Left / right hand sequences & $B \times T \times
  C_{hand}$ \\
  Conditioning & Object latent $z_{obj}$, object reconstruction
  $\hat{x}_{obj}$, PointNet features $f_{pointnet}$ & --- \\
  Preprocess & permute$(B,T,C_{hand}) \rightarrow
  (B,C_{hand},T)$ & $B \times C_{hand} \times T$ \\
  Shared encoder & Conv1d$(C_{hand} \rightarrow D, k{=}3, s{=}1,
  p{=}1)$ + ReLU & $B \times D \times T$ \\
  Encoder block & Conv1d$(D \rightarrow D, k{=}4, s{=}2, p{=}1)$
  + ResNet1D & $B \times D \times T/2$ \\
  Encoder block & Conv1d$(D \rightarrow D, k{=}4, s{=}2, p{=}1)$
  + ResNet1D & $B \times D \times T_{lat}$ \\
  Latent proj. & Conv1d$(D \rightarrow D, k{=}3, s{=}1, p{=}1)$
  & $B \times D \times T_{lat}$ \\
  Distribution mean & Conv1d$(D \rightarrow D, k{=}1)$ & $B
  \times D \times T_{lat}$ \\
  Distribution log-variance & Conv1d$(D \rightarrow D, k{=}1)$,
  clamp $[-10,10]$ & $B \times D \times T_{lat}$ \\
  Sampling & reparameterized sample $\rightarrow z_{lhand},
  z_{rhand}$ & $B \times D \times T_{lat}$ each \\
  Concat & concat$(z_{obj}, z_{lhand}, z_{rhand}, f_{pointnet})$
  & $B \times (3D{+}P) \times T_{lat}$ \\
  FiLM cond. & AvgPool1d$(k{=}s, \text{stride}{=}s)$ on
  $\hat{x}_{obj}$ $\rightarrow$ Conv1d$(C_{obj} \rightarrow
  2(3D{+}P), k{=}1)$ & $B \times 2(3D{+}P) \times T_{lat}$ \\
  FiLM split & split into $(\gamma, \beta)$ & $2 \times [B
  \times (3D{+}P) \times T_{lat}]$ \\
  FiLM apply & $\gamma \odot z + \beta$ & $B \times (3D{+}P)
  \times T_{lat}$ \\
  Decoder & Conv1d$(3D{+}P \rightarrow D)$ + ReLU & $B \times D
  \times T_{lat}$ \\
  Decoder block & ResNet1D $\rightarrow$ Upsample $\rightarrow$
  Conv1d & $B \times D \times T/2$ \\
  Decoder block & ResNet1D $\rightarrow$ Upsample $\rightarrow$
  Conv1d & $B \times D \times T$ \\
  Decoder & Conv1d$(D \rightarrow D)$ + ReLU & $B \times D
  \times T$ \\
  Output & Conv1d$(D \rightarrow 2C_{hand})$ & $B \times
  2C_{hand} \times T$ \\
  Postprocess & permute$(B,2C_{hand},T) \rightarrow
  (B,T,2C_{hand})$ & $B \times T \times 2C_{hand}$ \\
  \bottomrule
  \end{tabularx}
\caption{Network Architecture of the hand stream of the VAE.}
\label{tab:handae_arch}
\end{table}

\begin{table}[t]
\centering
\small
\setlength{\tabcolsep}{3pt}

\renewcommand{\arraystretch}{1.2}
\begin{tabular}{p{0.18\columnwidth} p{0.54\columnwidth} p{0.22\columnwidth}}
\hline
\textbf{Layer} & \textbf{Operation} & \textbf{Output Features} \\
\hline
\multicolumn{3}{l}{\textbf{A. Top-Level MARDM Pipeline}} \\
\hline
Input & Latent token sequence & $B \times L \times D_{ae}$ \\
InputProcess & $\mathrm{Linear}(D_{ae}\!\rightarrow\!D)$ & $B \times L \times D$ \\
Time + Modal Embedding & Add time and modality embeddings & $B \times L \times D$ \\
Transformer Stack & $\mathrm{MAR TransBlock} \times N_{tr}$ & $B \times L \times D$ \\
Diffusion Head & $\mathrm{DiffMLPs\_SiT}$ on masked tokens & $B \times L \times D_{ae}$ \\
\hline
\multicolumn{3}{l}{\textbf{B. MARTransBlock (Detailed)}} \\
\hline
AdaLN Modulation & $\mathrm{SiLU}\!\rightarrow\!\mathrm{Linear}(D\!\rightarrow\!6D)$, split to shifts/scales/gates & $6 \times (B \times D)$ \\
Multi-Head Self-Attention (MSA) & Q/K/V projections, scaled dot-product attention, output projection & $B \times L \times D$ \\
MLP & $\mathrm{Linear}(D\!\rightarrow\!D_{ff}\!\rightarrow\!D)$ with GELU & $B \times L \times D$ \\
Residual Gating & Gated residual updates for MSA and MLP branches & $B \times L \times D$ \\
\hline
\multicolumn{3}{l}{\textbf{C. DiffMLPs\_SiT (Detailed)}} \\
\hline
Input/Cond Proj & $\mathrm{Linear}(D_{ae}\!\rightarrow\!W_{diff})$, $\mathrm{Linear}(D\!\rightarrow\!W_{diff})$ & $N \times W_{diff}$ \\
Time Embedder & Sinusoidal timestep embedding + MLP & $N \times W_{diff}$ \\
Res Stack & $\mathrm{ResBlock} \times N_{diff}$ with adaLN gating & $N \times W_{diff}$ \\
Final Layer & AdaLN-modulated LN + $\mathrm{Linear}(W_{diff}\!\rightarrow\!D_{ae})$ & $N \times D_{ae}$ \\
Sampling & ODE sampler & $N \times D_{ae}$ \\
\hline
\end{tabular}%
\caption{Unified MAD-HOI generator architecture.}
\label{tab:gen_arch}
% \label{tab:mardm_unified_singlecol}
\end{table}

\subsection{Evaluation Pipeline}
\label[appendix]{app:eval_protocol}
We evaluate MAD-HOI with a learned text-motion evaluator adopted from HumanML3D \citep{guo2022generating}, following the standard practice of measuring generated motions in a semantic embedding space rather than directly in the raw motion parameter space. We train an evaluator with real hand-object interaction data to learn aligned representations for text and motion.

The evaluator consists of a motion representation module, a text encoder, and a motion encoder. On the motion side, we first compress the full hand-object motion sequence using a temporal convolutional movement encoder. This module reduces the temporal resolution and produces a compact latent motion representation that captures the essential dynamics of the sequence. A corresponding decoder is used only during pre-training of this motion representation. On the text side, we encode tokenized captions using pre-trained Word2Vec embeddings together with POS-style linguistic indicators, which are passed through a bidirectional GRU and projected into a shared embedding space. On the motion side, the compact motion latents are processed by a second bidirectional GRU and projected into the same embedding space. In the first stage of training, we learn the motion representation using an auto-encoding objective on real hand-object motion sequences. The model reconstructs the input sequence from its latent representation, while regularization terms encourage sparse and temporally smooth motion latents. This stage provides a compact and stable motion representation that is later used for text-motion alignment. In the second stage of training, we freeze the motion representation module and train the text encoder and motion encoder jointly with a contrastive objective. For each training sample, the caption and its paired motion are treated as a positive pair and are encouraged to have similar embeddings. Negative pairs are formed by mismatching captions and motions within the batch, so the evaluator learns to distinguish semantically correct interactions from incorrect ones. 

% The final evaluator therefore maps text and motion into a common semantic space where matched pairs are close, and mismatched pairs are far apart

% Given a text description, MAD-HOI produces a motion sequence for the hands and object interacting over time. For evaluation, each generated sequence is paired with its input text and compared with the corresponding ground-truth motion of the evaluation set. 

For text to motion evaluation, we generate motions from text prompts from MAD-HOI as well as baselines, and feed them to the frozen evaluator to compute all generation metrics. Both generated and real motions are mapped in the evaluators' motion embedding space, while captions are mapped in the corresponding text embedding space. We compute retrieval-style and distribution-level metrics in this shared space. 
% In particular, we report R-precision by measuring whether the correct motion is retrieved among the top nearest neighbors for a given text query, and we report a matching score based on the distance between paired text and motion embeddings. We also compute FID and KID on motion embeddings to measure how closely the generated motion distribution matches the ground-truth distribution, as well as diversity and multi-modality to quantify variety of generated motions. Additionally, we also report per-frame geometric metrics such as jitter, penetration, and contact percentage to assess smoothness and hand-object interaction quality in a 3D space.

\subsection{Metrics}
\label[appendix]{app:metrics}
\subsubsection{Generation/Distribution-based Metric Definitions} Top-3 Accuracy (R@3) measures how often the correct conditioning text is ranked among the three closest matches when retrieving in the shared embedding space between motions and conditions. Higher R@3 indicates higher text–motion semantic alignment. Frechet Inception Distance (FID) measures the discrepancy between the distributions of real and generated motion embeddings by comparing their means and covariance matrices under a Gaussian approximation. Kernel Inception Distance (KID) measures the discrepancy between real and generated motion embeddings using a kernel-based maximum mean discrepancy estimator. Unlike FID, it does not assume Gaussian feature distributions and is less biased for small sample sizes. Lower FID and KID indicate greater similarity between the generated and real motion distributions. Diversity is the variance within features extracted from the generated motions, and Matching Score is the embedding distance between generated motions and their conditioning text. Diversity values closer to the ground-truth diversity indicate more realistic coverage of the data distribution and lower Matching Scores indicate stronger text–motion semantic alignment. We omit Multimodality because its pairwise-distance formulation can reward unrestricted or implausible variation and lacks a corresponding ground-truth reference; diversity is instead assessed alongside distributional, semantic-alignment, physical-plausibility, and human-evaluation metrics. All our generation metrics are averaged over 5 evaluation repeats with varying seeds. The standard deviations were negligible and did not affect the ranking of methods; therefore, they are omitted for readability. The geometric metrics computed once only.  

\subsubsection{Completion and Infilling metrics:} 
Average Displacement Error (ADE) measures how closely the best generated motion matches the ground-truth motion over the entire sequence by averaging the displacement error across all time steps. Final Displacement Error (FDE) measures the displacement between the final pose of the ground-truth motion and the final pose of the closest generated sample. Lower ADE and FDE indicate better agreement with the ground-truth trajectory and endpoint, respectively, and are common metrics for evaluating sample accuracy in trajectory forecasting \citep{yuan2020dlow}. 

\subsubsection{EOM Metrics:} 
We evaluate End-of-Motion (EOM) prediction using both standard regression metrics and distribution-aware tolerance metrics. The mean absolute error (MAE) measures the average absolute difference between the predicted and ground-truth sequence length (number of frames), while the median absolute error (MedAE) captures the typical error and is less sensitive to outliers. 
%Root Mean Squared Error (RMSE) places greater weight on large mistakes, and Bias measures signed error, with values near zero indicating good calibration and positive or negative values indicating systematic over- or under-prediction. Pearson's r and Spearman's $\rho$ quantify linear and rank correlation, respectively, between predicted and true motion lengths. 
We also report Within-4 and Within-8 $\%$s, which denote the fraction of predictions that fall within tolerance thresholds of 4 and 8 frames from the ground truth sequence lengths, respectively. Since a single text prompt can correspond to multiple valid sequence lengths (from different human demonstrations) in the dataset, we additionally use distribution-aware metrics based on an interval $[q_{10}, q_{90}]$ for each unique condition (which is a [\text{action}, \text{object}, \text{left\_use}, \text{right\_use}] tuple). Interval MAE measures the distance to this valid range, assigning zero penalty when the prediction lies inside it, while Within Interval reports the proportion of predictions that fall within the semantically plausible duration range from the ground truth sequences.

In reference to \cref{tab:eom_metrics1}, MAD-HOI achieves strong and semantically meaningful EOM performance across both datasets, though it does not beat the Text2HOI cVAE. On ARCTIC, MAD-HOI attains an MAE of $12.23 \pm 0.88$ frames and a Median AE of 8.00, showing that its predicted stopping point is typically close to the true end of the motion. On GRAB where the problem is more challenging overall, MAD-HOI attains a higher MAE of $20.43 \pm 1.09$ frames and a Median AE of 14.4. 
Importantly, the distribution-aware metrics are particularly encouraging: an Interval MAE of $2.32 \pm 0.50$ for ARCTIC and $4.85 \pm 0.75$ for GRAB show that when we deviate from the distribution, we observe errors within 2-5 frames. The Within Interval \%s of $78.4\% \pm 1.3\%$ for ARCTIC $70.5\% \pm 1.3\%$ for GRAB and show that most MAD-HOI predictions fall inside the valid duration range for the corresponding semantic interaction, even when there is substantial natural variability in demonstration length.

\begin{figure}[t]
    \centering
    \includegraphics[width=\columnwidth]{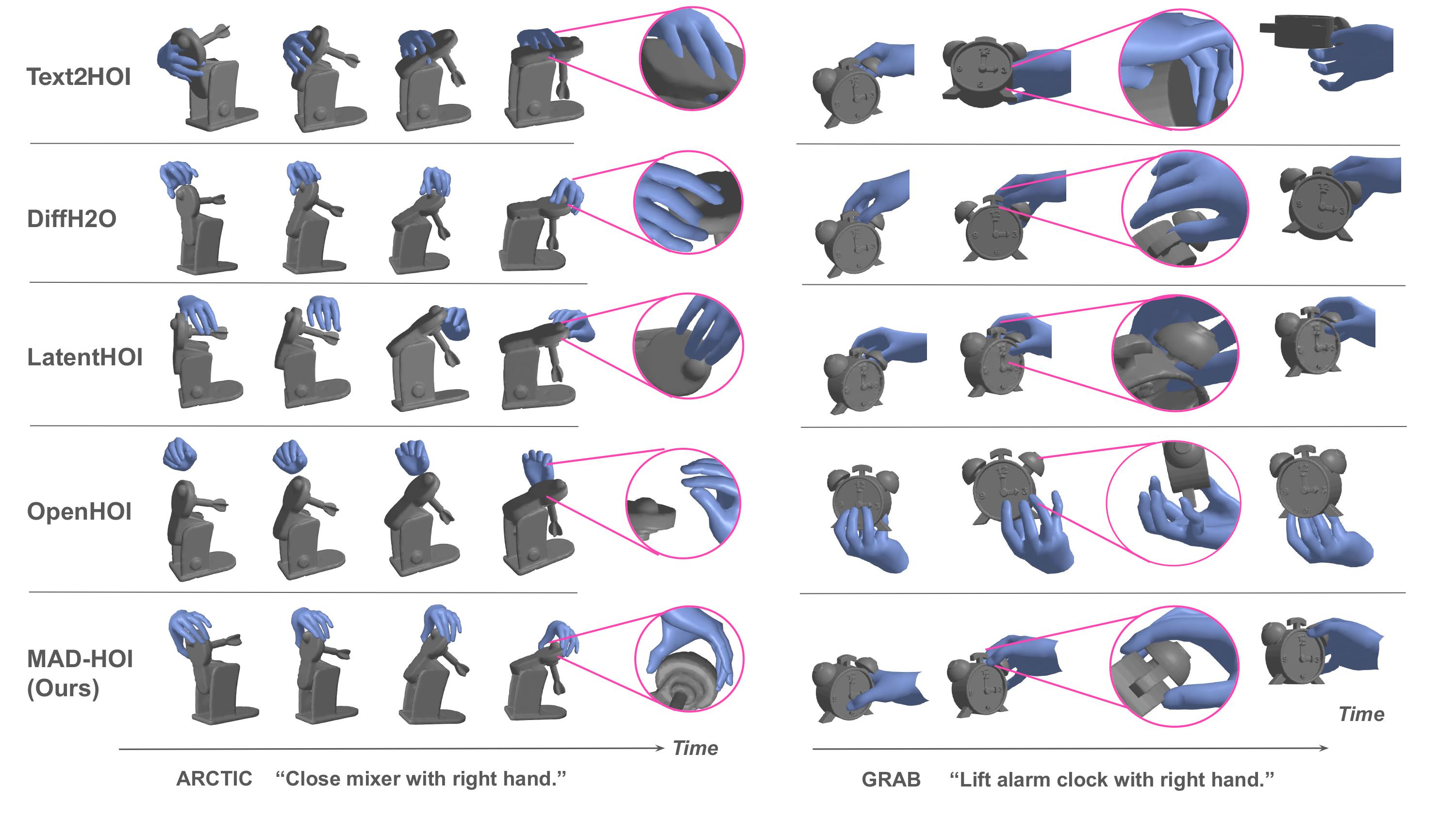}
    \caption{\textbf{Comparison of time-progressing grasps with baselines}. One frame is visualized with zoom-in to highlight contact and penetrations in generations across all methods. Right hand visualized in blue.}
    \label{fig:comp}
\end{figure}

\begin{figure}[t]
    \centering
    \includegraphics[width=0.98\columnwidth]{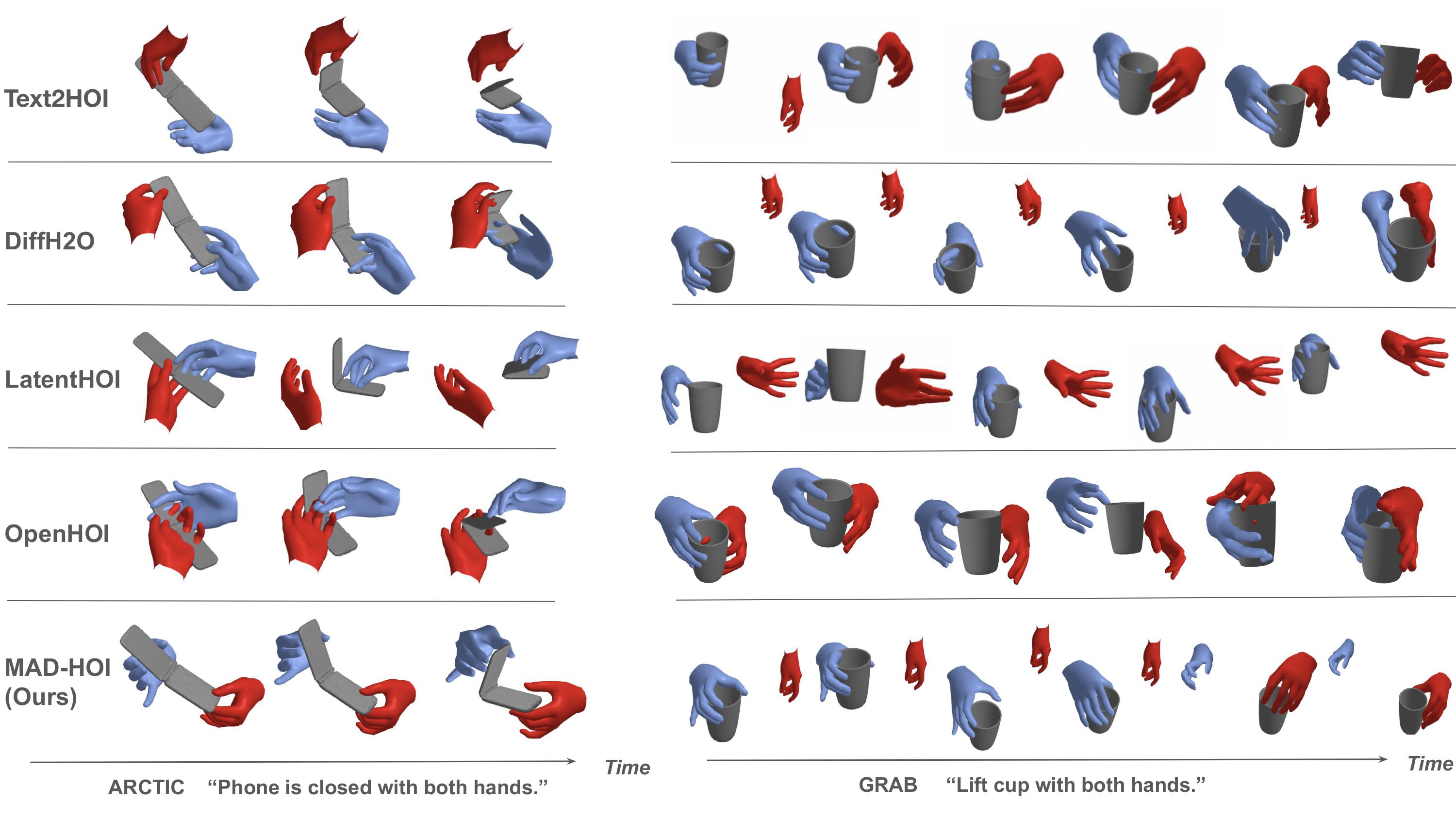}
    \caption{\textbf{Comparison of time-progressing grasps with baselines}. Right hand visualized in blue; Left hand visualized in red.}
    \label{fig:comp2}
\end{figure}

\subsection{Experiments}
We visualize sample motions from ARCTIC and GRAB datasets with time-progressing grasps for MAD-HOI and the baseline methods in \cref{fig:comp} and \cref{fig:comp2}. We observe that generations from our method align semantically better with the text prompts than baselines, both in terms of motion and hand use. We also observe that our method minimizes penetrations while maintaining consistent contact.  Please see supplementary material for interactive motion samples from our method and baselines.

\subsubsection{User Preference}
\label[appendix]{app:user_study}
We conduct a small user-study to evaluate human preferences for our method and the baselines. We evaluate the five 
  generation methods (MAD-HOI, DiffH2O, Text2HOI,
  OpenHOI, LatentHOI) through a forced-ranking study on
  ARCTIC (articulated objects) and GRAB (rigid
  objects). Each trial presents one text prompt
  alongside five anonymised, interactive 3D viewers
  labelled A–E, with the slot-to-method assignment
  randomised independently per trial so that no letter
  is associated with a method. Participants produce two
  complete 1–5 rankings per trial: motion–text 
  alignment — how faithfully the motion depicts the
  described action — and physical plausibility —
  naturalness of hand–object contact, penetration and
  object motion, judged independently of the text. All
  methods are rendered through an identical pipeline,
  with the same camera treatment, the same colour
  coding for left and right hands, and the same
  playback rate, so that no method is visually
  distinguishable from another. Each participant
  completes 50 trials, with prompts drawn
  without replacement from ARCTIC and GRAB
  prompts, and we collect three raters per dataset. 

  MAD-HOI attains the best mean rank on both
  criteria — 2.26 for motion–text alignment and 2.26 
  for physical plausibility (1 = best of five, 3.0 =
  chance) — ahead of DiffH2O (2.48 / 2.50), Text2HOI
  (2.81 / 2.91), OpenHOI (3.51 / 3.58) and LatentHOI
  (3.95 / 3.75), and is ranked first in 43.3\% of trials
  on both axes. Separation from Text2HOI, OpenHOI and
  LatentHOI is substantial, spanning 0.55 to 1.69 rank
  positions. DiffH2O is the closest competitor: in
  direct comparison MAD-HOI is preferred in 54.0% of
  trials for alignment and 55.3\% for plausibility. The
  study therefore indicates that MAD-HOI is preferred
  over three of the four baselines on both criteria,
  and is competitive with DiffH2O.

\subsubsection{Validating Handedness: Quantifying 'Unused Hand' Motion in MAD-HOI}
\label[appendix]{app:handedness}
In order to quantify the claim of our method learning to 'switch-off' unused hands, we present \cref{tab:idle_hand_comparison}. It evaluates the 'generated motion' for inactive hands in prompts specifying a single hand. The wrist and vertex displacement are in m, compounded over the entire sequence. The hand accuracy calculates the \% of frames where the unsued hand wrist (defined by prompt) moves less than 1 cm. Our idle latents corresponding to 'unused hands' are forced to predict zeros. And since neural networks learn smooth approximations, obtaining zero-latents for unused hands inevitably has some noise. This noise does lead to some non-zero cumulative joint motion over all the frames of a sequence, and therefore, the displacement errors are not zero. However,  the smaller inactive hand displacement MAD-HOI has compared to Text2HOI confirms that our method predicts near-zero motion for unused hands, and the presence of explicit latents for unused hands in our VAE. We also present a visual depiction of a frame from the generated motion for 'Close the waffleiron with right hand' from MAD-HOI. Here, the unused hand - the left hand - shown without any masking is truly inactive, while the right hand completes the desired motion. The generated poses for this hand are near-zeros, leading to the twisted MANO hand mesh. 

% \head{\textbf{
% % Idle-hand Validation \& User-specified Length:

% }}

% \vspace{2pt}

% \begin{center}
% \begin{minipage}{0.68\columnwidth}
% \centering

% \begin{minipage}[c]{0.16\linewidth}
%     \centering
%     \includegraphics[width=\linewidth]{images/idle_hand.png}
% \end{minipage}
% \hspace{0.03\linewidth}
% \begin{minipage}[c]{0.76\linewidth}
%     \centering
%     \scriptsize
%     \setlength{\tabcolsep}{2.5pt}
%     \renewcommand{\arraystretch}{0.92}

%     \begin{tabular}{l|cc|cc}
%     \toprule
%     \rowcolor{gray!25}
%     & \multicolumn{2}{c|}{\textbf{ARCTIC}}
%     & \multicolumn{2}{c}{\textbf{GRAB}} \\
%     \rowcolor{gray!15}
%     \textbf{Idle metric}
%     & \textbf{T2H}
%     & \textbf{Ours}
%     & \textbf{T2H}
%     & \textbf{Ours} \\
%     \midrule
%     Wrist disp.\ $\downarrow$
%     & 0.068 & \textbf{0.030}
%     & 0.274 & \textbf{0.003} \\
%     Vertex disp.\ $\downarrow$
%     & 0.078 & \textbf{0.033}
%     & 0.331 & \textbf{0.047} \\
%     Handedness Acc.\ $\uparrow$
%     & 0.54 & \textbf{0.64}
%     & 0.36 & \textbf{1.00} \\
%     \bottomrule
%     \end{tabular}
% \end{minipage}

% \end{minipage}
% \end{center}

% % \vspace{-2pt}
% \captionof{Table 10: Idle-hand comparison between Text2HOI and MAD-HOI.
% Inactive-hand wrist and vertex displacement and handedness accuracy on
% ARCTIC and GRAB.}
% \label[table]{tab:idle_hand_comparison}
% \vspace{6pt}

\begin{center}
\begin{minipage}{0.68\columnwidth}
\centering
\begin{minipage}[c]{0.16\linewidth}
    \centering
    \includegraphics[width=\linewidth]{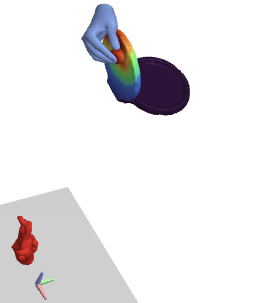}
\end{minipage}
\hspace{0.03\linewidth}
\begin{minipage}[c]{0.76\linewidth}
    \centering
    \scriptsize
    \setlength{\tabcolsep}{2.5pt}
    \renewcommand{\arraystretch}{0.92}
    \begin{tabular}{l|cc|cc}
    \toprule
    \rowcolor{gray!25}
    & \multicolumn{2}{c|}{\textbf{ARCTIC}}
    & \multicolumn{2}{c}{\textbf{GRAB}} \\
    \rowcolor{gray!15}
    \textbf{Idle metric}
    & \textbf{T2H} & \textbf{Ours}
    & \textbf{T2H} & \textbf{Ours} \\
    \midrule
    Wrist disp.\ $\downarrow$  & 0.068 & \textbf{0.030} & 0.274 & \textbf{0.003} \\
    Vertex disp.\ $\downarrow$ & 0.078 & \textbf{0.033} & 0.331 & \textbf{0.047} \\
    Handedness Acc.\ $\uparrow$ & 0.54 & \textbf{0.64} & 0.36 & \textbf{1.00} \\
    \bottomrule
    \end{tabular}
\end{minipage}
\end{minipage}
\captionof{table}{Idle-hand comparison between Text2HOI and MAD-HOI:
inactive-hand wrist and vertex displacement and handedness accuracy on
ARCTIC and GRAB.}
\label{tab:idle_hand_comparison}
\end{center}
\vspace{6pt}

Learning to independently model single-handed and bimanual motions without post-hoc masking is important when the interaction looks very different when being carried out with one hand vs both hands. For example, the grasp that a hand might make while opening a box might vary depending on whether a second hand stabilizes the box. Learning to differentiate single-handed motions from bimanual ones can create modes in the latent space which could be the difference between stable and plausible motions that actually achieve the desired object articulation. This is especially important when such text-to-motion models are applied to embodied use cases like robotics.

% {\small\textbf{(c)} Idle-hand metrics}

% \captionof{table}{Metrics for tokenizer design choice (a), Geometric Loss Ablations (b), and Idleness Verification (c).}
% \label{tab:ae_idle_rebuttal}

\subsubsection{Learned EOM vs User-Input Length} \label[appendix]{app:learned_eom}
We observed that predicting the EOM token helps MAD-HOI generate better motions. However, we argue that allowing the method to jointly estimate sequence length is better than using user-specified length. Our method, like our baselines, is not trained to generate motions at varying frame rates, and therefore, if a user input for sequence length is very different from lengths the model has seen during training, it could harm the quality of the generated motion. The assumption that the user has greater controllability by specifying sequence length hinges on the user being an expert and knowing the details of the training data. To demonstrate how 'wrong' length inputs can harm generation quality, we conduct an experiment where we force the learned EOM token to be at -4 frames, -8 frames and -12 frames from the ground truth lengths and compare the generated motions against the model-predicted EOM baseline. Forcing lengths greater than GT is not effective with our MAD-HOI baseline here as the method will predict an EOM token per its learned distribution and this will likely come before the user-specified EOM token. As seen in \cref{tab:eom_length_ablation}, we observe that the farther the user-specified length is from the ground truth length, the worse the quality of generated motion. The baseline performance - detected EOM in MAD-HOI - reflects the learned natural distribution of motions. When allowed to detect the sequence end, the model is able to generate more natural and expressive motions that are semantically aligned with the textual intent.

\begin{table*}[t] 
  \centering
  \small
  \setlength{\tabcolsep}{6pt}
  \renewcommand{\arraystretch}{1.05}
  
  \begin{tabular}{@{}l rrrrrrrr@{}}
  \toprule
  Length & FID $\downarrow$ & KID $\downarrow$ & Div. & Top-1 $\uparrow$ &
  Top-3 $\uparrow$ & Match $\downarrow$ & Pen.\,\% & Contact\,\% \\
  \midrule
  $-12$ frames & 4.240 & 0.0595 & 19.087 & 0.699 & 0.882 & 4.616 & 17.13 &
  79.89 \\
  $-8$ frames  & 0.623 & 0.0119 & 19.997 & 0.817 & 0.975 & 2.066 & 17.78 &
  81.37 \\
  $-4$ frames  & 0.130 & 0.0035 & 20.037 & 0.837 & 0.991 & 1.273 & 18.08 &
  81.58 \\
  \midrule
  MAD-HOI detected end & 0.075 & 0.0004 & 20.130 & 0.837 & 0.993 & 1.097 &
  15.25 & 83.86 \\
  \bottomrule
  \end{tabular}
  \caption{Forced EOM sequence-length ablation on ARCTIC. Motions are generated
  with the
  supplied length shortened by 4, 8, and 12 frames relative to the MAD-HOI
  detected
  end-of-motion. }
  \label{tab:eom_length_ablation}
  \end{table*}

\subsubsection{Baselines for Completion and Infilling} 
\label[appendix]{app:comp_inf_baseline}

 We compare MAD-HOI against simple, protocol-matched, non-learned kinematic baselines. For completion, we extrapolate the motion at the constant velocity given by the last two observed frames, extrapolating rotations on $\mathrm{SO}(3)$. For infilling, we interpolate between observed frames, using linear interpolation for translational and articulation components and SLERP (Spherical Linear Interpolation) for rotations; gaps observed on only one side hold the nearest observed frame, and sequences with no observed frame within their valid length fall back to the per-object mean pose computed from training sequences. For ADE/FDE, all methods are evaluated on the same test sequences, with the same masks and the same masked frames (for MAD-HOI, those within its predicted sequence length).

 The kinematic baselines provide a strong test of whether MAD-HOI learns interaction dynamics beyond simple local continuation. From \cref{tab:DE}, we see that for completion, MAD-HOI substantially outperforms constant-velocity extrapolation on GRAB across all ADE and FDE settings. On the shorter
ARCTIC sequences, constant-velocity extrapolation becomes competitive as more context is provided, reflecting the effectiveness of a local motion prior over short prediction horizons; nevertheless, MAD-HOI achieves fetter ADE and FDE when lesser context is provided and more completion is necessary (5 context tokens provided does shorten the number of frames left in the motion that need to be completed). The FID and Diversity scores also underline how MAD-HOI outperforms the constant velocity extrapolation.

For infilling, MAD-HOI consistently outperforms interpolation in ADE and FDE on
both ARCTIC and GRAB, while also achieving substantially lower FID. These results
show that the masked generative model does more than interpolate between observed
poses: it learns interaction dynamics that better reconstruct the missing motion
while remaining close to the distribution of valid HOI trajectories

 \begin{table}[H]
  \centering
  \scriptsize
  \setlength{\tabcolsep}{2.2pt}
  \renewcommand{\arraystretch}{1.05}
  \begin{adjustbox}{max width=\columnwidth}
  \begin{tabular}{@{}lll*{5}{c}!{\hspace{8pt}}*{5}{c}@{\hspace{5pt}}c@{\hspace{5pt}}c@{}}
  \toprule
  & & & \multicolumn{5}{c}{ADE $\downarrow$} & \multicolumn{5}{c}{FDE $\downarrow$} & FID
  $\downarrow$ & Diversity $\uparrow$ \\
  \cmidrule(lr){4-8}\cmidrule(lr){9-13}\cmidrule(lr){14-14}\cmidrule(lr){15-15}
  \multicolumn{3}{@{}l}{\textbf{Task Tokens (\#):}} & 1 & 2 & 3 & 4 & 5 & 1 & 2 & 3 & 4 &
  5 & 2 & 2 \\
  \midrule
  \multirow{4}{*}{Completion}
   & \multirow{2}{*}{ARCTIC}
     & Const.\ Vel. & 0.733 & 0.563 & 0.428 & 0.304 & 0.272 & 1.532 & 1.209 & 0.933 &
  0.656 & 0.589 & 3.873 & 18.823 \\
   & & MAD-HOI      & 0.649 & 0.617 & 0.613 & 0.608 & 0.592 & 0.816 & 0.787 & 0.789 &
  0.737 & 0.733 & 0.173 & 20.325 \\
  \cmidrule(l){2-15}
   & \multirow{2}{*}{GRAB}
     & Const.\ Vel. & 2.911 & 2.909 & 3.231 & 3.423 & 3.473 & 4.928 & 5.197 & 6.415 &
  7.260 & 7.612 & 14.546 & 11.174 \\
   & & MAD-HOI      & 0.748 & 0.776 & 0.738 & 0.749 & 0.724 & 0.963 & 1.019 & 1.001 &
  0.960 & 0.882 & 1.719 & 13.010 \\
  \midrule
  \multirow{4}{*}{Infilling}
   & \multirow{2}{*}{ARCTIC}
     & Interp. & 1.284 & 1.135 & 1.026 & 0.953 & 0.869 & 1.325 & 1.158 & 1.041 &
  0.951 & 0.856 & 7.025 & 18.582 \\
   & & MAD-HOI           & 0.783 & 0.707 & 0.675 & 0.639 & 0.629 & 0.833 & 0.766 & 0.719 &
  0.691 & 0.684 & 0.182 & 19.214 \\
  \cmidrule(l){2-15}
   & \multirow{2}{*}{GRAB}
     & Interp. & 1.224 & 0.917 & 0.764 & 0.651 & 0.545 & 1.337 & 1.209 & 1.077 &
  0.942 & 0.842 & 4.582 & 12.707 \\
   & & MAD-HOI           & 0.721 & 0.579 & 0.546 & 0.514 & 0.484 & 0.844 & 0.671 & 0.614 &
  0.591 & 0.536 & 1.931 & 13.649 \\
  \bottomrule
  \end{tabular}
  \end{adjustbox}
  \vspace{-2pt}
  \caption{Conditioned completion and infilling against non-learned kinematic baselines. ADE/FDE are computed on the predicted frames over hand-joint trajectories (21 joints per active hand) and the object position trajectory, each z-scored with dataset statistics and averaged with equal weight (unitless). Const.\ Vel.: constant-velocity extrapolation from the last two observed frames (rotations on SO(3)) till GT sequence length. Interp.: linear/SLERP interpolation between observed frames. Lower ADE, FDE, and FID are better; higher Diversity is better.}
  \label{tab:DE}
  \vspace{-10pt}
  \end{table}

\subsubsection{Generalization}
\label[appendix]{app:general}
In order to test the robustness of MAD-HOI, we design another split for the ARCTIC dataset where we exclude certain motion samples with single handedness (ex. \textit{'Close Laptop with left hand', 'Right hand opens mixer'} from the training set and evaluate the trained model on these excluded samples to test generalization of motion transfer between hands. Our base ARCTIC split is termed \textbf{\textit{'Level 1'}} and the split with these exclusions is termed \textbf{\textit{'Level 2'}}. We provide metrics for text to motion on both splits in \cref{tab:arctic_splits_metrics}. As the performance of our model on the Level 2 ARCTIC split is not much worse than Level 1, we conclude that our modality tokens help the model generalize to out-of-distribution motions regarding \textit{'handedness'}. 

When we tested our method for similar splits which exclude certain 'actions' or 'objects', we did not see robust generalization and this is expected as the \textit{'action'} signal is primarily encoded by CLIP embeddings and the \textit{'object geometry'} is encoded by pointnet features only. Therefore, for more intent-based or geometry-based generalization, we require richer descriptors and perhaps training on much larger datasets which can offer a diverse grasp prior.

\begin{table}[t]
\centering

\setlength{\tabcolsep}{8pt}
\small
\resizebox{\textwidth}{!}{%
\begin{tabular}{l c c c c c c c c}
\toprule
\textbf{Split} & \begin{tabular}[c]{@{}c@{}}\textbf{Accuracy}\\\textbf{(Top 3)} $\uparrow$\end{tabular} & \textbf{FID} $\downarrow$ & \textbf{KID} $\downarrow$ & \textbf{Diversity} $\rightarrow$ & \textbf{Matching} $\downarrow$ & \begin{tabular}[c]{@{}c@{}}\textbf{Accel.}\\\textbf{(Jitter)} $\downarrow$\end{tabular} & \begin{tabular}[c]{@{}c@{}}\textbf{Penetration}\\\textbf{(\%)} $\downarrow$\end{tabular} & \begin{tabular}[c]{@{}c@{}}\textbf{Contact}\\\textbf{(\%)} $\uparrow$\end{tabular} \\
\midrule
\cellcolor{gray!15}Ground Truth & \cellcolor{gray!15}0.991  & \cellcolor{gray!15}- & \cellcolor{gray!15}- & \cellcolor{gray!15}20.235 & \cellcolor{gray!15}1.295 & \cellcolor{gray!15}0.187 & \cellcolor{gray!15}7.864  & \cellcolor{gray!15}80.771\\
Level 1 & 0.944  & 0.071 & 0.004 & 20.312 & 1.353 & 0.18 & 12.89  & 84.47 \\
Level 2 & 0.914 & 0.137 & 0.006 & 20.717 & 1.525 & 0.299 & 13.012 & 80.637 \\
\bottomrule
\end{tabular}%
}
\caption{Quantitative evaluation of text to motion for Level 1 and Level 2 splits from ARCTIC. Ground truth values are provided for reference and are highlighted in gray.}
\label{tab:arctic_splits_metrics}
\end{table}

\subsection{Application}
\label[appendix]{app:application_robot}
High-fidelity 4D HOI sequences can serve as priors for robotic motion by capturing detailed human-object interactions.
These sequences provide smooth trajectories with realistic contact events and limited penetration between bodies, properties that matter when learning control policies.
Furthermore, text-driven generation of these sequences reduces the dependence on manually collected human demonstrations, which otherwise require extensive and labor-intensive teleoperation setups \citep{khazatsky2024droid, o2024open, qiu2025humanoid}.

% \begin{figure}[!htp]
%   \centering
%   \includegraphics[scale=0.22]{images/applications_inspire.png}
%   \caption{\textbf{Robot retargeting from MAD-HOI outputs.}
%   A text-generated HOI sequence is retargeted from the MANO model to the Unitree~G1 robot with Inspire dexterous hands.
%   The resulting trajectory preserves realistic contact and grasp transitions, indicating that our generated motions can serve as reference priors for robotic manipulation.
%   }
%   \label{fig:applications_inspire}
% \end{figure}

% When used as reference motion, they guide reinforcement learning toward desirable regions of the action space, improving convergence and reducing the likelihood of physically impossible behaviors.

% Moreover, by supplying rich supervisory signals that would otherwise require extensive teleoperation or motion-capture efforts, our approach substantially reduces the dependence on manually collected human demonstrations. This enables scalable policy learning and broadens the range of tasks that can be addressed without expensive data collection pipelines.

Although an embodiment gap exists between the MANO model and dexterous robot hands, we find that our generated sequences can seed a manipulation pipeline.
We bridge this gap in multiple stages: first, the MANO motion is retargeted to the robot hand using keypose-vector optimization \citep{qin2023anyteleop}.
This target motion is then translated to the full robot using Inverse Kinematics.
Finally, this reference trajectory is refined using the physics-based rewards of DexMachina's \citep{mandi2025dexmachina} functional retargeting pipeline to produce a fine-tuned policy.
Fig. \ref{fig:applications_inspire} shows a robot trajectory, with motion retargeted from our generated sequence to the Unitree G1 robot with Inspire hands.

\subsection{Future Work}
While we verify generalization across motion handedness, generalizing to unseen objects and actions will require pretraining on large datasets. Whether this pretraining would be more effective with static grasps (much higher scale of data available but lacks temporal context) or short motion priors (temporal data but scarce), is unknown and can be studied in more detail. Perhaps a two-stage generation setup with the first stage focusing on static grasps and the second stage focusing on motion could work well. However, as seen in our baselines that adopt this type of architecture, such a design is better suited for motions and objects that do not require articulations (DiffH2O and LatentHOI work better on GRAB than ARCTIC).

\clearpage
\renewcommand{\refname}{Appendix References}
\putbib
\end{bibunit}
\end{document}